\documentclass{article}

\usepackage[preprint]{neurips_2021}

\usepackage[utf8]{inputenc} %
\usepackage[T1]{fontenc}    %
\usepackage[pagebackref=true, colorlinks=true,urlcolor=blue,citecolor=black,linkcolor=blue,anchorcolor=black]{hyperref}       %
\usepackage{url}            %
\usepackage{booktabs}       %
\usepackage{amsfonts}       %
\usepackage{nicefrac}       %
\usepackage{microtype}      %
\usepackage{xcolor}         %

\usepackage{graphicx}
\usepackage{amsmath}
\usepackage{amssymb}
\usepackage{xspace}
\usepackage{xurl}
\usepackage{enumitem}
\usepackage{placeins}
\usepackage{capt-of}
\usepackage{array}
\usepackage[ruled,vlined]{algorithm2e}

\usepackage{cleveref}
\usepackage{wrapfig}
\usepackage{threeparttable}
\usepackage[disable]{todonotes}
\usepackage[skip=4pt]{caption}
\usepackage{enumitem}
\usepackage{subcaption}

\newcommand{\model}{SGI\xspace}

\crefname{paragraph}{paragraph}{paragraphs}

\title{Pretraining Representations for Data-Efficient Reinforcement Learning}

\author{Max Schwarzer\thanks{\{max.schwarzer, nitarshan\}@mila.quebec}$^{*1,2}$, Nitarshan Rajkumar$^{*1,2}$, Michael Noukhovitch$^{1,2}$, Ankesh Anand$^{1,2}$\\
\vspace{-3mm}
\\\textbf{Laurent Charlin$^{1,3,5}$, Devon Hjelm$^{1,4}$, Philip Bachman$^{1,4}$, Aaron Courville$^{1,2,5}$}\\
\vspace{-2mm}
\\$^1$Mila, $^2$Université de Montréal, $^3$HEC Montréal, $^4$Microsoft Research, $^5$CIFAR
}

\begin{document}

\maketitle

\begin{abstract}
Data efficiency is a key challenge for deep reinforcement learning.
We address this problem by using unlabeled data to pretrain an encoder which is then finetuned on a small amount of task-specific data.
To encourage learning representations which capture diverse aspects of the underlying MDP, we employ a combination of latent dynamics modelling and unsupervised goal-conditioned RL.
When limited to 100k steps of interaction on Atari games (equivalent to two hours of human experience), our approach significantly surpasses prior work combining offline representation pretraining with task-specific finetuning, and compares favourably with other pretraining methods that require orders of magnitude more data.
Our approach shows particular promise when combined with larger models as well as more diverse, task-aligned observational data -- approaching human-level performance and data-efficiency on Atari in our best setting.
We provide code associated with this work at \url{https://github.com/mila-iqia/SGI}.

\end{abstract}
\section{Introduction}

Deep reinforcement learning (RL) methods often focus on training networks \textit{tabula rasa} from random initializations without using any prior knowledge about the environment. In contrast, humans rely a great deal on visual and dynamics priors about the world to perform decision making efficiently \citep{dubey2018investigating, lake2017building}. Thus, it is not surprising that RL algorithms which learn \textit{tabula rasa} suffer from severe overfitting~\citep{zhang2018study} and poor sample efficiency compared to humans~\citep{tsividis2017-learning91}. 

Self-supervised learning (SSL) provides a promising approach to learning useful priors from past data or experiences.
SSL methods leverage unlabelled data to learn strong representations, which can be used to bootstrap learning on downstream tasks.
Pretraining with self-supervised learning has been shown to be quite successful in vision~\citep{cpcv2, BYOL, simCLR} and language~\citep{devlin2019-bert,gpt3} settings. 

Pretraining can also be used in an RL context to learn priors over representations or dynamics. One approach to pretraining for RL is to train agents to explore their environment in an unsupervised fashion, forcing them to learn useful skills and representations~\citep{Hansen2020Fast,liu2021unsupervised,campos2021coverage}. %
Unfortunately, current unsupervised exploration methods require months or years of real-time experience, which 
may be impractical for real-world systems with limits and costs to interaction --- agents cannot be run faster than real-time, may require significant human oversight for safety, and can be expensive to run in parallel.
It is thus important to develop pretraining methods that work with practical quantities of data, and ideally that can be applied \textit{offline} to fixed datasets collected from prior experiments or expert demonstrations~\citep[as in][]{stooke2020decoupling}.

To this end, we propose to use a combination of self-supervised objectives for representation learning on offline data, requiring orders of magnitude %
less pretraining data than existing methods, while approaching human-level data-efficiency when finetuned on downstream tasks.
We summarize our work below:
\vspace{-5pt}
\paragraph{RL-aligned representation learning objectives:}
We propose to pretrain representations using a combination of latent dynamics modeling, unsupervised goal-conditioned reinforcement learning, and inverse dynamics modeling -- with the intuition that a collection of such objectives can capture more information about the dynamical and temporal aspects of the environment of interest than any single objective.
We refer to our method as \textbf{SGI} (\textbf{S}PR, \textbf{G}oal-conditioned RL and \textbf{I}nverse modeling).
\vspace{-5pt}
\paragraph{Significant advances for data-efficiency on Atari:}
\model's combination of objectives performs better than any in isolation and significantly improves performance over previous representation pretraining baselines such as ATC~\citep{stooke2020decoupling}.
Our results are competitive with exploration-based approaches such as APT or VISR~\citep{liu2021unsupervised, Hansen2020Fast}, which require two to three orders of magnitude more pretraining data and the ability to interact with the environment during training, while SGI can learn with a small offline dataset of exploration data.
\vspace{-5pt}
\paragraph{Scaling with data quality and model size:}
\model's performance also scales with data quality and quantity, increasing as data comes from better-performing or more-exploratory policies.
Additionally, we find that \model's performance scales robustly with model size; while larger models are unstable or bring limited benefits in standard RL, \model pretraining allows their finetuning performance to significantly exceed that of smaller networks.

We assume familiarity with RL in the following sections (with a brief overview in~\Cref{app:rl_background}).

\section{Representation Learning Objectives}

\begin{figure}[t]
    \centering
    \includegraphics[width=0.99\textwidth]{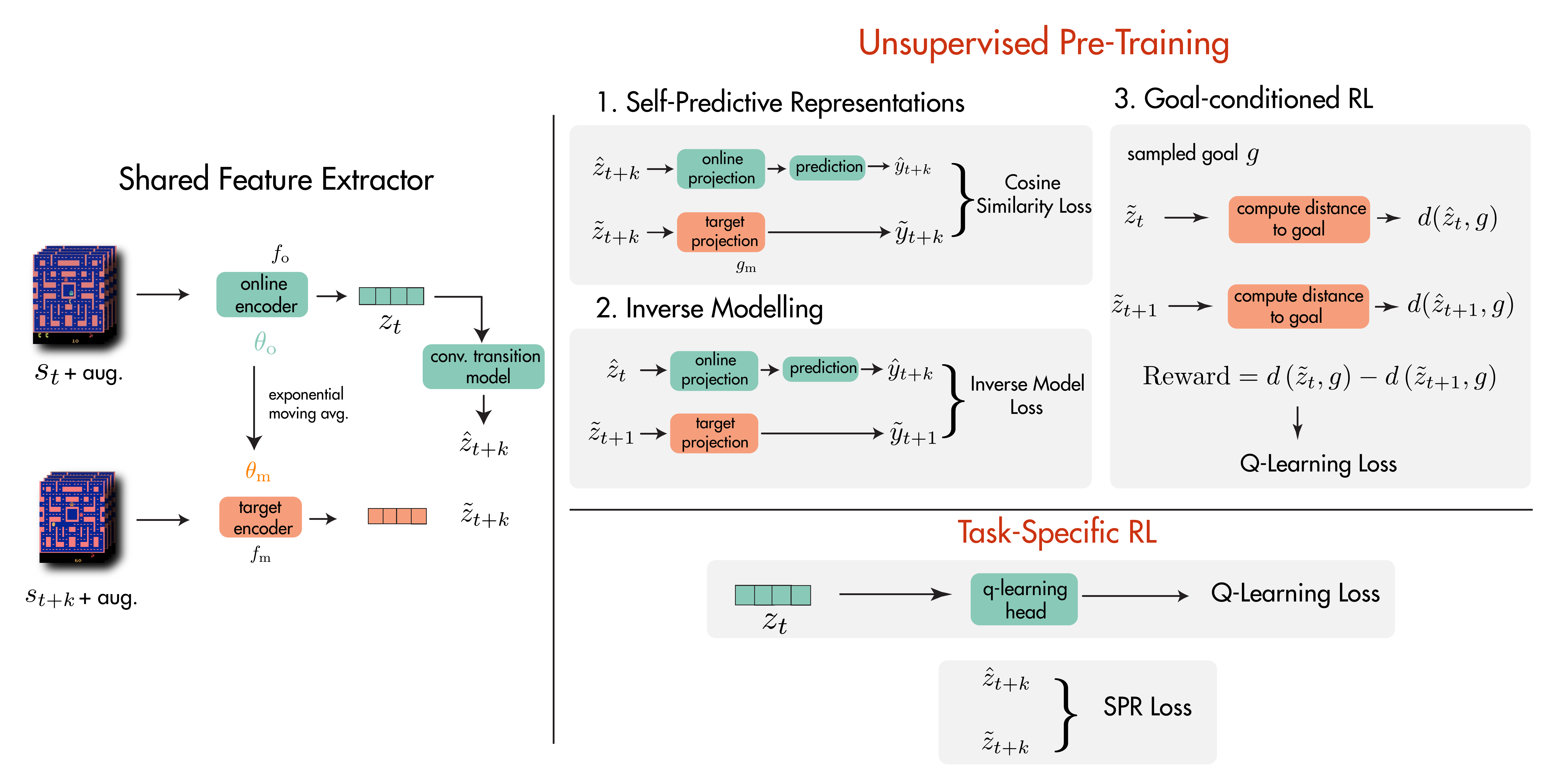}
    \caption{
    A schematic diagram showing our two stage pretrain-then-finetune method.
    All unsupervised training losses and task-specific RL use the shared torso on the left.
    }
    \label{fig:diagram}
    \vspace{-\intextsep}
\end{figure}

A wide range of SSL objectives have been proposed for RL which leverage various aspects of the structure available in agent interactions. For example, the temporal dynamics of an environment can be exploited to create a forward prediction task~\citep[e.g.,][]{deepMDP, guo2018neural, stooke2020decoupling,  schwarzer2020dataefficient} in which an agent is trained to predict its immediate future observations, perhaps conditioned on a sequence of actions to perform.

Structure in RL goes far beyond forward dynamics, however.
We propose to combine multiple representation learning objectives, covering different agent-centric and temporal aspects of the MDP.
Based on the intuition that knowledge of an environment is best represented in multiple ways~\citep{hessel2021muesli, degris2012scaling}, we expect this to outperform monolithic representation learning methods such as temporal contrastive learning~\citep[e.g.,][]{stooke2020decoupling}.
In deciding which tasks to use, we consider questions an adequate representation should be able to answer about its environment, including:

\begin{itemize}[topsep=0pt,itemsep=0pt,partopsep=0pt, parsep=0pt]
    \item If I take action $a$ in state $s$, what state $s'$ am I likely to see next?
    \item If I transitioned from state $s$ to state $s'$, what action $a$ did I take?
    \item What action $a$ would I take in state $s$ so that I reach another state $s'$ as soon as possible
\end{itemize}

Note that none of these questions are tied to task reward, allowing them to be answered in a fully-unsupervised fashion.
Additionally, these are questions about the environment, and not any specific policy, allowing them to be used in offline pretraining with arbitrary behavioral policies.

In general, the first question may be answered by forward dynamics modeling, which as mentioned above is well-established in RL.  The second question corresponds to inverse dynamics modeling, which has also been commonly used in the past~\citep{lesort2018state}.
The third question corresponds to self-supervised goal-conditioned reinforcement learning which has the advantage of being structurally similar to the downstream target task, as both require learning to control the environment.

To facilitate their joint use, we formulate these objectives so that they operate in the latent representation space provided by a shared encoder.
We provide an overview of these components in \Cref{fig:diagram} and describe them in greater detail below; we also provide detailed pseudocode in~\Cref{pseudocode}.

\subsection{Self-Predictive Representations}
SPR~\citep{schwarzer2020dataefficient} is a representation learning algorithm developed for data-efficient reinforcement learning.
SPR learns a latent-space transition model, directly predicting representations of future states without reconstruction or negative samples.
As in its base algorithm, Rainbow~\citep{hessel2018rainbow}, SPR learns a convolutional encoder, denoted as $f_o$, which produces representations of states as $z_t = f_o(s_t)$.
SPR then uses a \emph{dynamics model} $h$ to recursively estimate the representations of future states, as $\hat{z}_{t+k+1} = h(\hat{z}_{t+k}, a_{t+k})$, beginning from $\hat{z_t} \triangleq z_{t}$.
These representations are projected to a lower-dimensional space by a projection function $p_o$ to produce $\hat{y}_{t+k} \triangleq p_o(\hat{z}_{t+k})$.

Simultaneously, SPR uses a \emph{target encoder} $f_m$ to produce target representations $\tilde{z}_{t+k} \triangleq f_m(s_{t+k})$, which are further projected by a target projection function $p_m$ to produce $\tilde{y}_{t+k} \triangleq p_m(\tilde{z}_{t+k})$.
SPR then maximizes the cosine similarity between these predictions and targets, using a learned linear prediction function $q$ to translate from $\hat{y}$ to $\tilde{y}$:
\begin{align}{
    \mathcal{L}_\theta^{\text{SPR}}(s_{t:t+K}, a_{t:t+K}) = -\sum_{k=1}^K  \frac{q(\hat{y}_{t+k}) \cdot {\tilde{y}}_{t+k} }{||q(\hat{y}_{t+k})||_2 \cdot  || \tilde{y}_{t+k}||_2}.
}
\end{align} 
The parameters of these target modules $\theta_m$ are defined as an exponential moving average of the parameters $\theta_o$ of $f_o$ and $p_o$: $\theta_m = \tau \theta_m + (1 - \tau)\theta_o$.

\subsection{Goal-Conditioned Reinforcement Learning}
Inspired by works such as~\citet{dabney2020value} that show that modeling many different value functions is a useful representation learning objective, we propose to augment SPR with an unsupervised goal-conditioned reinforcement learning objective.
We define goals $g$ to be normalized vectors of the same size as the output of the agent's convolutional encoder (3,136- or 4,704-dimensional vectors, for the architectures we consider).
We use these goals to annotate transitions with synthetic rewards, and train a modified version of Rainbow~\citep{hessel2018rainbow} to estimate $Q(s_t, a, g)$, the expected return from taking action $a$ in state $s_t$ to reach goal $g$ if optimal actions are taken in subsequent states.

We select goals using a scheme inspired by hindsight experience replay~\citep{andrychowicz2017hindsight}, seeking to generate goal vectors that are both semantically meaningful and highly diverse.  As in hindsight experience replay, we begin by sampling a state from another trajectory or the future of the current trajectory.  However, we take the additional step of applying stochastic noise to encourage goals to lie somewhat off of the current representation manifold. We provide details in \Cref{app:gcrl}.

\subsection{Inverse Dynamics Modeling}
We propose to use an inverse dynamics modeling task~\citep{lesort2018state}, in which the model is trained to predict $a_t$ from $s_t$ and $s_{t+1}$.
Because this is a classification task (in discrete control) or regression task (continuous control), it is naturally not prone to representational collapse, which may complement and stabilize our other objectives.
We directly integrate inverse modeling into the rollout structure of SPR, modeling $p(a_{t+k}|\hat{y}_{t+k}, \tilde{y}_{t+k+1})$ for each $k \in \left(0, \ldots, {K}{-}{1} \right)$, using a two-layer MLP trained by cross-entropy.

\section{Related Work}

\paragraph{Data-Efficiency}
In order to address data efficiency in RL, \citet{simple} introduced the Atari 100k benchmark, in which agents are limited to 100,000 steps of environment interaction, and proposed SimPLe, a model-based algorithm that substantially outperformed previous model-free methods.
However, \citet{der} and \citet{kielak2020do} found that simply modifying the hyperparameters of existing model-free algorithms allowed them to exceed SimPLe's performance.
Later, DrQ~\citep{drq} found that adding mild image augmentation to model-free methods dramatically enhanced their sample efficiency, while SPR~\citep{schwarzer2020dataefficient} combined data augmentation with an auxiliary self-supervised learning objective.
\model employs SPR as one of its objectives in offline pretraining, leading to significant improvements in data-efficiency.

\paragraph{Exploratory pretraining}
A number of recent works have sought to improve reinforcement learning via the addition of an unsupervised \emph{pretraining} stage prior to finetuning on the target task.
One common approach has been to allow the agent a period of fully-unsupervised interaction with the environment, during which the agent is trained to maximize a surrogate exploration-based task such as the diversity of the states it encounters, as in APT~\citep{liu2021unsupervised}, or to learn a set of skills associated with different paths through the environment, as in DIAYN~\citep{eysenbach2018diversity}, VISR~\citep{Hansen2020Fast}, and DADS~\citep{sharma2019dynamics}.
Others have proposed to use self-supervised objectives to generate intrinsic rewards encouraging agents to visit new states; e.g. \citet{pathak2017curiosity} and \citet{burda2018largescale} use the loss of an inverse dynamics model like that used in \model, while \citet{sekar2020planning} uses the disagreement between an ensemble of latent-space dynamics models.
Finally, \citet{campos2021coverage} report strong results based on massive-scale unsupervised pretraining.

Many of these methods are used to pretrain agents that are later adapted to specific reinforcement learning tasks.
However, \model differs in that it can be used offline and is agnostic to how data is collected.
As such, if no pre-existing offline data is available, one of the methods above can be used to generate a dataset for \model; we use \citet{burda2018largescale} for this in \Cref{sec:dataset}.%

\paragraph{Visual Representation Learning}
Computer vision has seen a series of dramatic advances in self-supervised representation learning, including contrastive methods~\citep{cpc, dim, amdim, moco, simCLR} as well as purely predictive ones~\citep{BYOL}.
Variants of these approaches have also been shown to improve performance when coupled with a small quantity of labeled data, in a \emph{semi-supervised} setting~\citep{simCLRv2, cpcv2}, and several self-supervised methods have been designed specifically for this case~\citep[for example,][]{sohn2020fixmatch, tarvainen2017mean}.

These advances have spurred similar growth in methods aimed specifically at improving performance in RL.
We refer the reader to~\citet{lesort2018state} for a review of earlier methods, including inverse dynamics modeling which is used in \model.
Recent research has focused on leveraging latent-space dynamics modeling as an auxiliary task.
\citet{deepMDP} propose a simple next-step prediction task, coupled with reward prediction, but found it is prone to latent space collapse and requires an auxiliary reconstruction loss for experiments on Atari.
\citet{pbl} use a pair of networks for both forward and backward prediction, and show improved performance in extremely large-data multi-task settings.
\citet{mazoure2020deep} use a temporal contrastive objective for representation learning, and show improvement in continual RL settings.
Concurrently, SPR~\citep{schwarzer2020dataefficient} proposed a multi-step latent prediction task with similarities to BYOL~\citep{BYOL}.

Two works similar to ours, \citet{stdim} and \citet{stooke2020decoupling}, propose reward-free temporal-contrastive methods to pretrain representations.
\citet{stdim} show that representations from encoders trained with ST-DIM contain a great deal of information about environment states, but they do not examine whether or not representations learned via their method are, in fact, useful for reinforcement learning.  However, \citet{stooke2020decoupling} employ a similar algorithm and find only relatively minor improvements in performance compared to standard baselines in the large-data regime; our controlled comparisons show that SGI's representations are far better for data-efficiency. Concurrent to our work, FERM \citep{zhan2020framework} propose contrastive pretraining from human demonstrations in a robotics setting.  As FERM is quite similar to ATC, we are optimistic that our improvements over ATC in Atari 100k would translate to FERM's setting.

\section{Experimental Details}
In our experiments, We seek to address two main challenges for the deployment of RL agents in the real world~\citep{dulacarnold2020realworldrlempirical}: (1) training the RL agent with a limited budget of interactions in the real environment, and (2) leveraging existing interaction data of \textbf{arbitrary quality}.

\subsection{Environment and Evaluation}
To address the first challenge, we focus our experimentation on the Atari 100k benchmark introduced by~\citet{simple}, in which agents are allowed only 100k steps of interaction with their environment.\footnote{Note that sticky actions are disabled under this benchmark.}
This is roughly equivalent to the two hours human testers were given to learn these games by~\citet{dqn}, providing a baseline of human sample-efficiency.

Atari is also an ideal setting due to its complex observational spaces and diverse tasks, with 26 different games included in the Atari 100k benchmark.
These factors have led to Atari's extensive use for representation learning and exploratory pretraining~\citep{stdim, stooke2020decoupling, campos2021coverage}, and specifically Atari 100k for data-efficient RL~\citep[e.g.,][]{simple, drq, schwarzer2020dataefficient}, including finetuning after exploratory pretraining~\citep[e.g.,][]{Hansen2020Fast, liu2021unsupervised}, providing strong baselines to compare to.

Our evaluation metric for an agent on a game is \emph{human-normalized score} (HNS), defined as $\frac{agent\_score - random\_score}{human\_score - random\_score}$.
We calculate this per game by averaging scores over 100 evaluation trajectories at the end of training, and across 10 random seeds for training.
We report both mean (Mn) and median (Mdn) HNS over the 26 Atari-100K games, as well as on how many games a method achieves super-human performance ($>$H) and greater than random performance ($>$0).  Following the guidelines of~\citet{rl_precipice} we also report interquartile mean HNS (IQM) and quantify uncertainty via bootstrapping in~\Cref{app:uncertainty}. %

\subsection{Pretraining Data}
\label{sec:dataset}
The second challenge pertains to pretraining data.  Although some prior work on offline representational pretraining has focused on expert-quality data~\citep{stooke2020decoupling}, we expect real-world pretraining data to be of greatly varying quality.  We thus construct four different pretraining datasets to approximate different data quality scenarios.

\begin{itemize}[itemsep=2pt,leftmargin=1.15pc]
    \item \textbf{(R)andom} To assess performance near the lower limit of data quality, we use a random policy to gather a dataset of 6M transitions for each game.  To encourage the agent to venture further from the starting state, we execute each action for a random number of steps sampled from a Geometric$(\frac{1}{3})$ distribution.
    \item \textbf{(E)xploratory} To emulate slightly better data that covers a larger range of the state space, we use an exploratory policy. Specifically, we employ the \textbf{IDF} (inverse dynamics) variant of the algorithm proposed by~\cite{burda2018largescale}. We log the first 6M steps from an agent trained in each game.  This algorithm achieves better-than-random performance on only 70\% of tasks, setting it far below the performance of more modern unsupervised exploration methods.
\end{itemize}
To create higher-quality datasets, we follow~\citet{stooke2020decoupling} and use experience gathered during the training of standard DQN agents~\citep{dqn}.  We opt to use the publicly-available DQN Replay dataset~\citep{agarwal2020optimistic}, which contains data from training for 50M steps (across all 57 games, with five different random seeds).
Although we might prefer to use data from recent unsupervised exploration methods such as APT~\citep{liu2021unsupervised}, VISR~\citep{Hansen2020Fast}, or CPT~\citep{campos2021coverage}, none of these works provide code or datasets, making this impractical.
We address using data collected from on-task agents with a behavioural cloning baseline in \Cref{sec:comparison}, with surprising findings relative to prior work.

\begin{itemize}[itemsep=2pt,leftmargin=1.15pc]
    \item \textbf{(W)eak} We first generate a weak dataset by selecting the first 1M steps for each of the 5 available runs in the DQN Replay dataset.  This data is generated with an $\epsilon$-greedy policy with high, gradually decaying $\epsilon$, leading to substantial action diversity and many suboptimal exploratory actions.
    Although the behavioral policies used to generate this agent are not especially competent (see Table~\ref{tab:pretrain_data}), they have above-random performance on almost all games, suggesting that that this dataset includes more task-relevant transitions.
    
    \item \textbf{(M)ixed}
    Finally, for a realistic best-case scenario, we create a dataset of both medium and low-quality data. To simulate a real-world collection of data from different policies, we concatenate multiple checkpoints evenly spread throughout training of a DQN.
    We believe this is also a reasonable approximation for data from a modern unsupervised exploration method such as CPT \citep{campos2021coverage};
    as shown in Table~\ref{tab:pretrain_data}, the agent for this dataset has performance in between CPT and VISR, with median closer to CPT and mean closer to VISR.
    This data is also lower quality than the expert data originally used in the method most similar to ours, ATC~\citep{stooke2020decoupling}.~\footnote{Our data-collection agents are weaker than those used by ATC on seven of the eight games they consider.} We create a dataset of 3M steps and a larger dataset of 6M steps; all \textbf{M} experiments use the 3M step dataset unless otherwise noted.
\end{itemize}

\begin{wraptable}[19]{R}{7.7cm}
\vspace{-\intextsep}
\scalebox{1.}{
\begin{threeparttable}
\caption{Performance of agents used in pretraining data collection compared to external baselines on 26 Atari games \citep{simple}}
\label{tab:pretrain_data}
\begin{tabular}{lrrrrr}
\toprule
Method & Mdn & Mean & $>$H & $>$0 & Data \\
\midrule
\multicolumn{6}{l}{\textit{Exploratory Pretraining Baselines}} \\
\midrule
VISR@0 & 0.056 & 0.817 & 5 & 19 & 250M \\
APT@0\tnote{1} & 0.038 & 0.476 & 2 & 18 & 250M\\
CPT@0 & \textbf{0.809} & \textbf{4.945}  & \textbf{12} & 25 & 4B\\
\midrule
\multicolumn{6}{l}{ \textit{Offline Datasets}} \\
\midrule
Exploratory & 0.039 & 0.042 & 0 & 18 & 6M \\
Weak\tnote{2} & 0.028 & 0.056 & 0 & 23    & 5M \\
Mixed\tnote{2} & 0.618 & 1.266 & 10 & \textbf{26} & 3M \\
\bottomrule
\end{tabular}
\begin{tablenotes}%
\item[1] Calculated from ICLR 2021 OpenReview submission; unreported in arXiv version.
\item[2] Upper-bound estimate from averaging evaluation performance of corresponding agents in Dopamine.
\end{tablenotes}
\end{threeparttable}
}
\end{wraptable}

We compare the agents used for our datasets to those for unsupervised exploration pretraining baselines in Table~\ref{tab:pretrain_data}. We estimate the performance of the Weak and Mixed agents as the average of the corresponding logged evaluations in the Dopamine~\citep{castro18dopamine} baselines. Even our largest dataset is quite small compared to the amounts of data gathered by unsupervised exploration methods (see the ``Data'' column in Table \ref{tab:pretrain_data}); this is intentional, as we believe that unsupervised interactional data may be expensive in real world applications.  We show the performance of the non-random data collection policies in~\Cref{tab:HNS} (note that a fully-random policy has a score of \textbf{0} by definition).

\subsection{Training Details}

We optimize our three representation learning objectives jointly during unsupervised pretraining, summing their losses. 
During finetuning, we optimize only the reinforcement learning and forward dynamics losses, following \citet{schwarzer2020dataefficient} (see \Cref{sec:sgi_finetuning}), and lower the learning rates for the pretrained encoder and dynamics model by two orders of magnitude (see \Cref{sec:learning_rates}).

We consider the standard three-layer convolutional encoder introduced by~\citet{dqn}, a ResNet inspired by~\citet{impala}, as well as an enlarged ResNet of the same design.  In other respects, our implementation matches that of SPR and is based on its \href{https://github.com/mila-iqia/spr}{publicly-released code}.  Full implementation and hyperparameter details are provided in \Cref{app:implementation}.
We refer to agents by the model architecture and pretraining dataset type used: \textbf{\model-R} is pretrained on Random, \textbf{\model-E} on Exploratory, \textbf{\model-W} on Weak, and \textbf{\model-M} on Mixed.
To investigate scaling, we vary the size of the encoder used in \textbf{\model-M}: the standard \citet{dqn} encoder is \textbf{\model-M/S} (for small), our standard ResNet is simply \textbf{\model-M} and using a larger ResNet is \textbf{\model-M/L} (for large)\footnote{See \Cref{app:implementation} for details on these networks}.  
For \textbf{SGI-M/L} we also use the 6M step dataset described earlier.
All ablations are conducted in comparison to \textbf{SGI-M} unless otherwise noted.
Finally, agents without pretraining are denoted \textbf{SGI-None}; SGI-None/S would be roughly equivalent to SPR~\citep{schwarzer2020dataefficient}.

For baselines, we compare to no-pretraining Atari 100k methods \citep{simple,der,drq,schwarzer2020dataefficient}. For our models trained on Random and Exploratory data we compare against previous pretraining-via-exploration approaches applied to Atari 100k \citep{liu2021unsupervised, Hansen2020Fast, campos2021coverage}.
In the higher quality data regime, we compare to recent work on data-agnostic unsupervised pretraining, ATC \citep{stooke2020decoupling}, as well as  behavioural cloning (BC).%

\section{Results and Discussion}
\label{sec:discussion}
\begin{wraptable}[27]{R}{0.5\textwidth}
    \vspace{-2\intextsep}
    \label{tab:results_hns}
    \centering
    \scalebox{.90}{
    \begin{threeparttable}
    \caption{
        HNS on Atari100k for SGI and baselines.
    }\label{tab:HNS}
    \begin{tabular}{lrrrrr}
    \toprule
    Method & Mdn & Mn & $>$H & $>$0 & Data \\
    \midrule
    \multicolumn{6}{l}{\textit{No Pretraining (Finetuning Only)}} \\
    \midrule
    SimPLe & 0.144 & 0.443 & 2 & \textbf{26} &  0  \\
    DER & 0.161 & 0.285 & 2 & \textbf{26} & 0  \\
    DrQ & 0.268 & 0.357 & 2 & 24 & 0  \\
    SPR & 0.415 & 0.704 & 7 & \textbf{26} & 0 \\
    \textcolor{red}{SGI-None} & 0.343 & 0.565 & 3 & 26 & 0 \\
    \midrule
    \multicolumn{6}{l}{\textit{Exploratory Pretraining + Finetuning}} \\
    \midrule
    VISR & 0.095 & 1.281 & 7 & 21 & 250M \\
    APT & 0.475 & 0.666\tnote{1} & 7 & \textbf{26} & 250M\\
    CPT@0\tnote{2} & \textbf{0.809} & \textbf{4.945}  & 12 & 25 & 4000M \\
    \midrule
    ATC-R\tnote{3} & 0.191 & 0.472 & 4 & \textbf{26} & 3M  \\
    ATC-E\tnote{3} & 0.237 & 0.462 & 3 & \textbf{26} & 3M  \\
    \textcolor{red}{SGI-R} & 0.326 & 0.888 & 5 & \textbf{26} & 6M  \\
    \textcolor{red}{SGI-E} & 0.456 & 0.838 & 6 & \textbf{26} & 6M  \\
    \midrule
    \multicolumn{6}{l}{ \textit{Offline-data Pretraining + Finetuning}} \\
    \midrule
    ATC-W\tnote{3} & 0.219 & 0.587 & 4 & \textbf{26} & 3M  \\
    ATC-M\tnote{3} & 0.204 & 0.780 & 5 & \textbf{26} & 3M  \\
    BC-M@0 & 0.139 & 0.227 & 0 & 23 & 3M \\
    BC-M & 0.548 & 0.858 & 8 & \textbf{26} & 3M \\
    \textcolor{red}{SGI-W} & 0.589 & 1.144 & 8 & \textbf{26} & 5M  \\
    \textcolor{red}{SGI-M/S} & 0.423 & 0.914 & 8 & \textbf{26} & 3M  \\
    \textcolor{red}{SGI-M} & 0.679 & 1.149 & \textbf{9} & \textbf{26} & 3M  \\
    \textcolor{red}{SGI-M/L} & \textbf{0.753} & \textbf{1.598} & \textbf{9} & \textbf{26} & 6M  \\ 
    \bottomrule
    \end{tabular}
    \begin{tablenotes}[para]
    \item[1] APT claims 0.6955, but we calculate 0.666 from their reported per-game scores.
    \item[2] CPT@0 does not do any finetuning.
    \item[3] Our implementation (see \Cref{app:atc})
    \end{tablenotes}
    \end{threeparttable}
    }
\end{wraptable}

We find that \model performs competitively on the Atari-100K benchmark; presenting aggregate results in~\Cref{tab:HNS}, and full per-game data in~\Cref{app:full_results}.
Our best setting, \textbf{\model-M/L}, achieves a median HNS of 0.753, approaching human-level sample-efficiency and outperforming all comparable methods except the recently proposed CPT~\citep{campos2021coverage}.
With less data and a smaller model, \textbf{\model-M} achieves a median HNS of 0.679, significantly outperforming the prior method ATC on the same data (\textbf{ATC-M}).
Meanwhile, \textbf{SGI-E} achieves a median HNS of 0.456, matching or exceeding other exploratory methods such as APT~\citep{liu2021unsupervised} and VISR~\citep{Hansen2020Fast}, as well as ATC-E.

\paragraph{Pretraining data efficiency}
SGI achieves strong performance with only limited pretraining data; our largest dataset contains 6M transitions, or roughly 4.5 days of experience.  This compares favourably to recent works on unsupervised exploration such as APT or CPT, which require far larger amounts of data and environment interaction (250M steps or 193 days for APT, 4B steps or 8.45 years for CPT).  We expect SGI would perform even better if used in these large-data settingss, as we find that it scales robustly with data  (see \Cref{sec:scaling_models}).

\begin{wrapfigure}[17]{r}{0.5\textwidth}
    \includegraphics[width=0.5\textwidth]{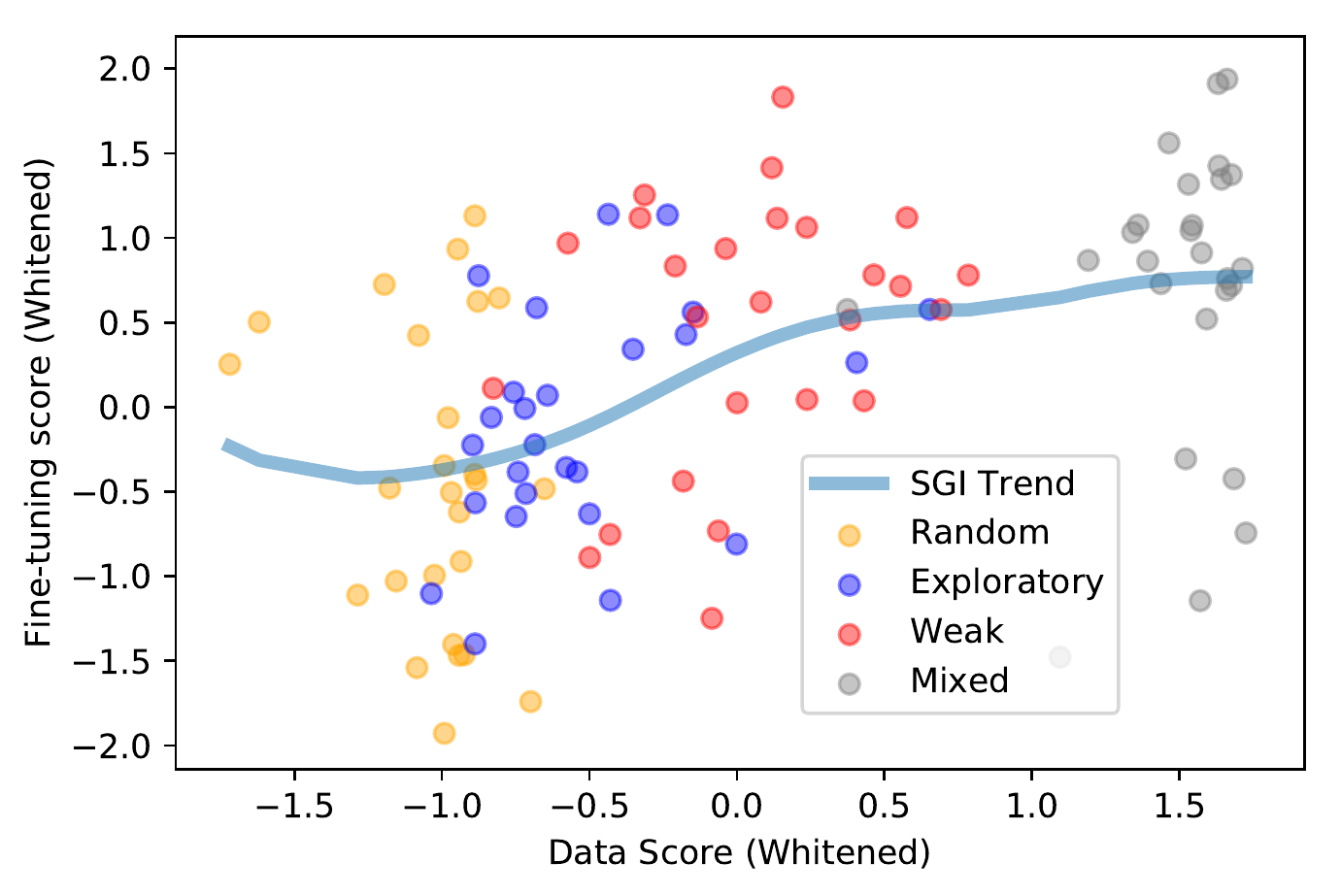}
    \caption{SGI finetuning performance vs. pretraining data score for all combinations of game and dataset.  Data score is estimated as clipped return per episode, trend calculated via kernel regression. Values whitened per-game for clarity.}
    \label{fig:data_quality}
\end{wrapfigure}

\paragraph{Behavioural cloning is a strong baseline}\label{sec:comparison}
Although ATC pretrains with expert data, they did not investigate behavioral cloning as a baseline pretraining objective.
We do so on our \textbf{Mixed} dataset, the only one to be generated by policies with significantly above-random performance.
Behavioral cloning without finetuning (\textbf{BC-M@0}) performs poorly, perhaps due to the varying behavioural quality in the dataset. But when finetuned, \textbf{BC-M} yields very respectable performance, surpassing \textbf{ATC-M} but not \textbf{SGI-M}.  All fine-tuning settings for BC-M match \model-M.

\subsection{Data quality matters}
In principle, \model can be used with any offline dataset but we demonstrate that it scales with the quality of its data.
Near the lower bound of data quality where all actions are selected randomly, \textbf{SGI-R} still provides some benefit over an otherwise-identical randomly-initialized agent (\textbf{SGI-None}) on 16 out of 26 games, with a similar median but 57\% higher mean HNS. With better data from an exploratory policy, \textbf{SGI-E} improves on 16/26 games, gets 33\% higher median HNS, and surpasses APT~\citep{liu2021unsupervised} which used 40 times more pretraining data. With similarly weak data but possibly more task-specific transitions, \textbf{\model-W} gets 72\% higher median HNS compared to \model-None and with realistic data from a mixture of policies, \textbf{\model-M} improves to 98\%.

Importantly, the pattern we observe is very different from what would be expected for imitation learning.  In particular, SGI-W's strong performance shows that expert data is not required.  To characterize this, we plot the average clipped reward~\footnote{Unclipped rewards are not available for the offline DQN dataset.} experienced per episode for each of our pretraining datasets in \Cref{fig:data_quality}. Normalizing across tasks, we find a strong positive correlation between task reward engagement ($p<0.0001$) and finetuning performance.  Moreover, we find diminishing returns to further task engagement.

\subsection{Pretraining unlocks the value of larger networks}
\label{sec:scaling_models}

\begin{figure}[t]
\begin{minipage}{0.48\textwidth}
\centering
\includegraphics[width=\linewidth]{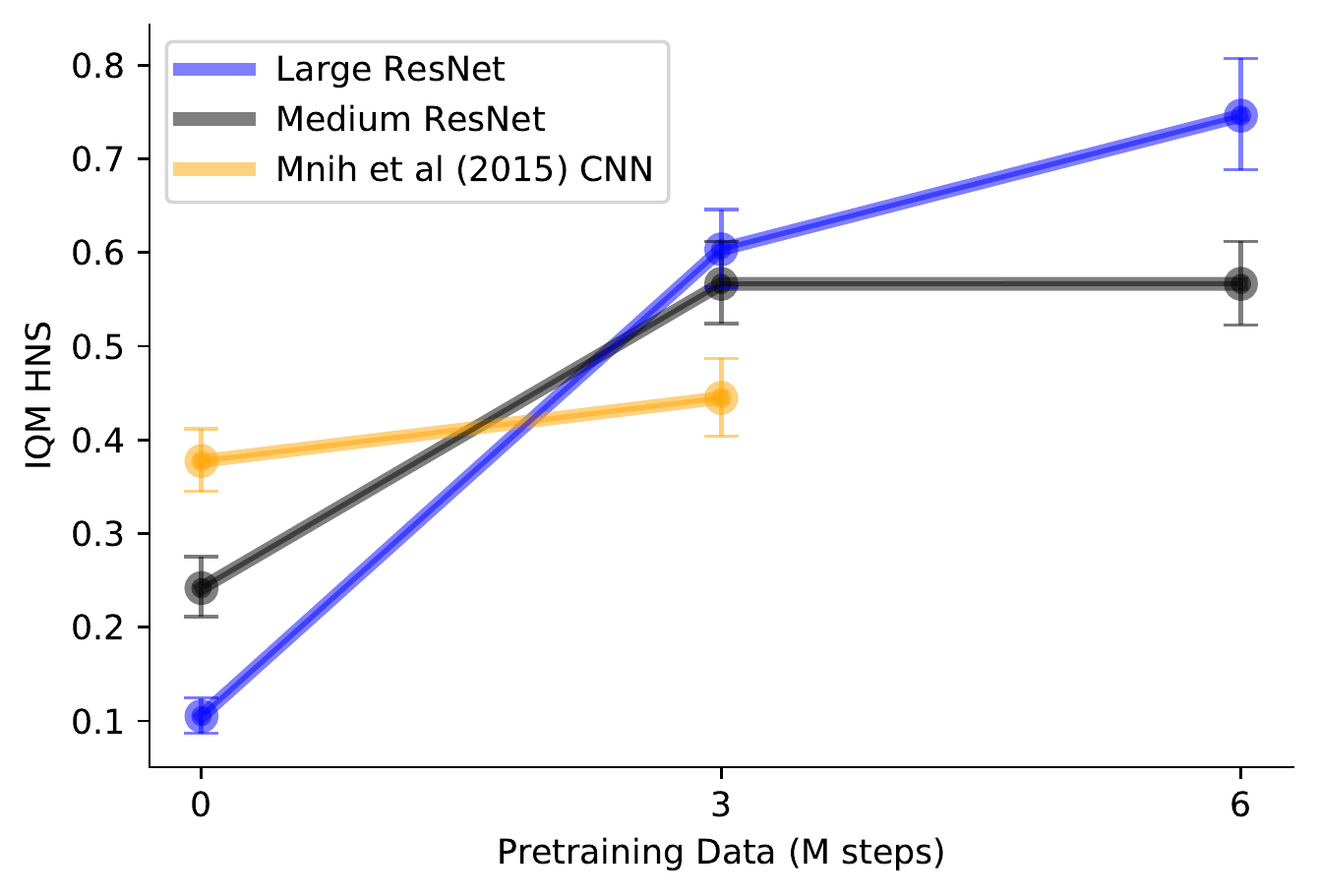}
\caption{
Finetuning performance of \model for different CNN sizes and amounts of pretrained data from the \textbf{Mixed} dataset.  We plot IQM HNS with confidence intervals (see \Cref{app:uncertainty}).
}
\label{fig:data_size}
\end{minipage}\hfill
\begin{minipage}{0.49\textwidth}
\centering
\includegraphics[width=\linewidth]{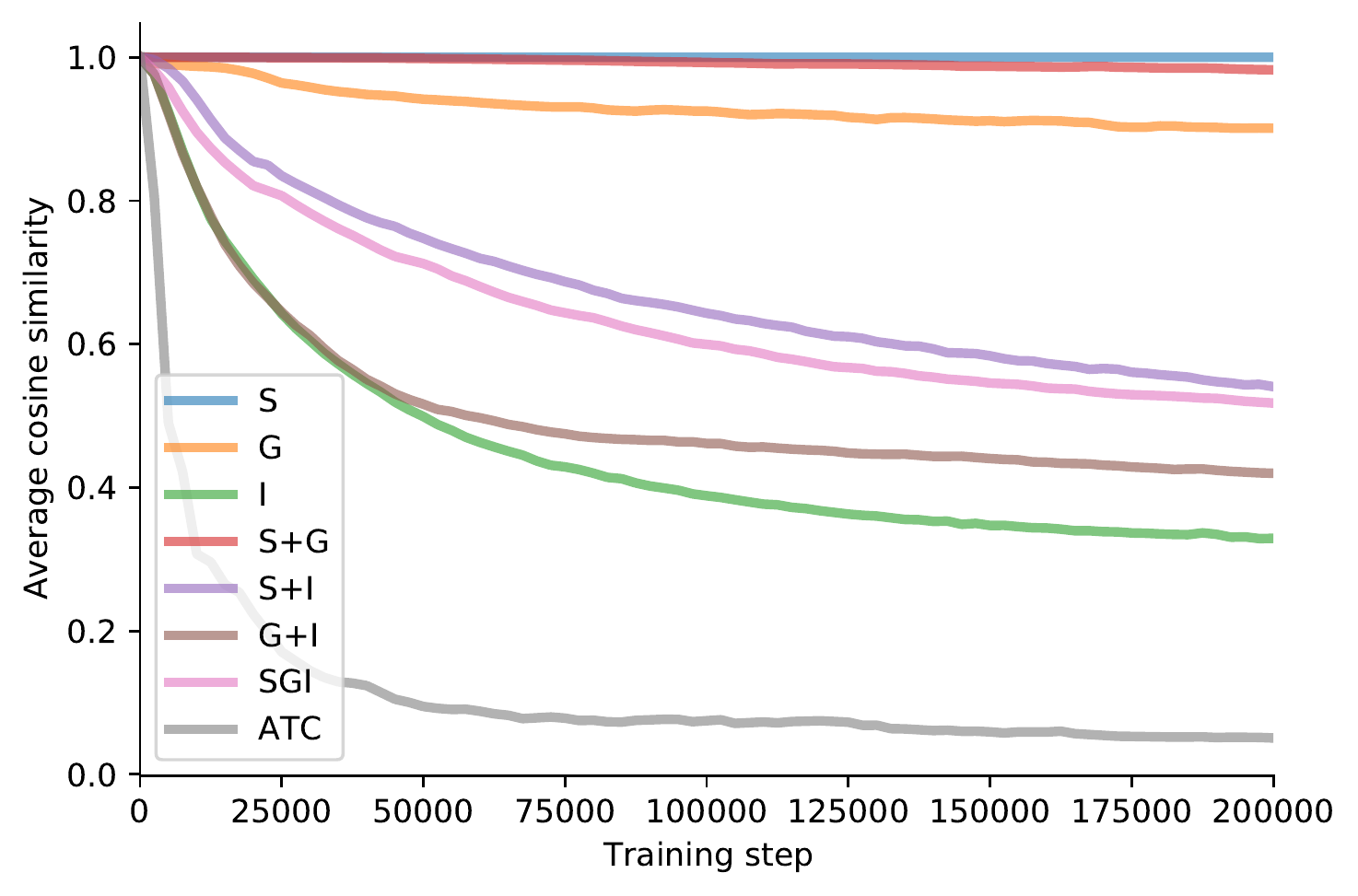}
\caption{
    Average cosine similarity between representations over pretraining, averaged across the 26 Atari 100k games.
    1 indicates representations are identical, 0 perfect dissimilarity.
}\label{fig:collapse}
\end{minipage}
\vspace{-\intextsep}
\end{figure}

The three-layer network introduced by~\citet{dqn} has become a fixture of deep reinforcement learning, and has been employed by previous works examining pretraining in this field~\citep[e.g.][]{liu2021unsupervised, stooke2020decoupling}.
However, we find that representational pretraining with this network (\textbf{SGI-M/S}) provides only minor benefits compared to training from scratch. In contrast, larger networks struggle without pretraining but shine when pretrained as shown in~\Cref{fig:data_size}.

This finding is consistent with recent work in unsupervised representation learning for classification, which has observed that unsupervised pretraining benefits disproportionately from larger networks~\citep{simCLR}.
In particular, our results suggest that model size should increase in parallel with amount of pretraining data, matching recent work on scaling in language modeling~\citep{kaplan2020scaling, hernandez2021scaling}.
\model thus provides a simple way to use unlabeled data to extend the benefits of large networks, already well-known in the large-data regime~\citep[e.g.,][]{muzero, impala}, to data-efficient RL.

\begin{wraptable}[12]{r}{6cm}
\vspace*{-1.4\intextsep}
\scalebox{1}{
\begin{threeparttable}
\caption{
    HNS on Atari 100K for pretraining ablations of SGI.
}\label{tab:controls_hns}
\begin{tabular}{lrrr}
\toprule
Pretraining & Mdn & Mean & $>$H \\
\midrule
None & 0.343 & 0.565 & 3 \\
S & 0.009 & -0.054 & 0 \\
G & 0.060 & 0.181 & 1 \\
I & 0.411 & 0.943 & 7 \\
S+G & 0.029 & 0.098 & 0 \\
G+I & 0.512 & 1.004 & \textbf{9} \\
S+I & 0.629 & 0.978 & 8 \\
\midrule
SGI-M & \textbf{0.679} & \textbf{1.149} & \textbf{9} \\
\bottomrule
\end{tabular}
\end{threeparttable}
}
\end{wraptable}
\subsection{Combining SGI's objectives improves performance}
We test all possible combinations of our three SSL objectives, denoted by combinations of the letters S, G and I to indicate which objectives they employ.
Results in \Cref{tab:controls_hns} show that performance monotonically increases as more objectives are used, with inverse dynamics modeling combined with either of the other objectives performing respectably well. This illustrates the importance of using multiple objectives to obtain representational coverage of the MDP.

We note that including inverse modeling appears to be critical, and hypothesize that this is related to representational collapse.
To measure this, we plot the average cosine similarity between representations $y_t$ of different states for several pretraining methods in \Cref{fig:collapse}, using our ResNet encoder on the Mixed dataset. We observe that \textbf{S}, \textbf{G} and \textbf{S+G} all show some degree representational collapse, while configurations that include inverse dynamics modeling avoid representational collapse, as does ATC, whose contrastive loss implicitly optimizes for representational diversity~\citep{wang2020understanding}.
Intriguingly, we observe that increased representational diversity does not necessarily improve performance.  For example, SGI outperforms \textbf{ATC}, \textbf{G+I} and \textbf{I} in finetuning but has less diverse pretraining representations.
We also observe that adding SPR (\textbf{S}) consistently pulls representations towards collapse (compare \textbf{S+I} and \textbf{I}, \textbf{S+G} and \textbf{G}, and \textbf{SGI} and \textbf{G+I}); how this relates to performance is a question for future work.

\subsection{Naively finetuning ruins pretrained representations}\label{sec:learning_rates}

\begin{wraptable}[9]{R}{5.85cm}
\vspace*{-\intextsep}
\centering
\scalebox{1.0}{
\begin{threeparttable}

\caption{
    HNS on Atari 100K for fine-tuning schemes for SGI.
}\label{tab:finetune_controls_hns}

\begin{tabular}{lrrr}
\toprule
Method  & Mdn   & Mean  & $>$H \\
\midrule
No pretrain & 0.343 & 0.565 & 3\\
Naive   & 0.429 & 0.845 & 8    \\
Frozen  & 0.499 & 0.971 & 8    \\
Reduced LRs & \textbf{0.679} & \textbf{1.149} & \textbf{9} \\ %
\bottomrule
\end{tabular}

\end{threeparttable}
}
\end{wraptable}
We find that properly finetuning pretrained representations is critical, as results in \Cref{tab:finetune_controls_hns} show. Although allowing pretrained weights to change freely during finetuning is better than initializing from scratch (\textbf{Naive} vs \textbf{No Pretrain}), freezing the pretrained encoder (\textbf{Frozen}) leads to better performance than either.  \model's approach of reducing finetuning learning rates for pretrained parameters leads to superior performance (\textbf{Reduced LRs}, equivalent to \textbf{SGI-M}).

We thus hypothesize that representations learned by \model are being disrupted by gradients early in finetuning, in a phenomenon analogous to catastrophic forgetting~\citep{zenke2017continual, hadsell2020embracing}.  As representations may not generalize between different value functions across training~\citep{dabney2020value}, allowing the encoder to strongly adapt early in training could make it \emph{worse} at modeling later value functions, compared to the neutral initialization from \model.  We also note that there is a long history in computer vision of employing specialized finetuning hyperparameters~\citep{Li2020Rethinking, chu2016best} when transferring tasks.

\subsection{Not all SSL objectives are beneficial during finetuning}
\label{sec:sgi_finetuning}
\begin{wraptable}[14]{R}{5.8cm}
\vspace*{-\intextsep}
\caption{
    HNS on Atari 100K for finetuning ablations of SGI.
}\label{tab:ssl_ft}
\centering
\scalebox{1.}{
\begin{tabular}{lrrrr}
\toprule
Finetune SSL & Mdn & Mean & $>$H \\
\midrule
\multicolumn{4}{l}{\textit{Without SGI pretraining}} \\
\midrule
None & 0.161 & 0.315 & 2\\
S Only & 0.343 & 0.565 & 3\\
\midrule
\multicolumn{4}{l}{\textit{With SGI pretraining}} \\
\midrule
None & 0.452 & 1.114 & 8 \\
SGI & 0.397 & 1.011 & 8 \\
S Only & \textbf{0.679} & \textbf{1.149} & \textbf{9} \\
\bottomrule
\end{tabular}
}
\end{wraptable}
Although \model uses \textbf{S} during finetuning, we experiment with a variant that optimizes only the standard DQN objective, roughly equivalent to using DrQ~\citep{drq} with DQN hyperparameters set to match \model.  We find that finetuning with \textbf{S} has a large impact with or without pretraining (compare \textbf{None} and \textbf{S Only} entries in \Cref{tab:ssl_ft}.).  Although, \model without \textbf{S} manages to achieve roughly the same mean human-normalized score as \model with \textbf{S}, removing \textbf{S} harms performance on performance on 19 out of 26 games and reduces median normalized score by roughly 33\%.  We also find no benefit to using all of \model's constituent objectives during finetuning (\textbf{All Losses} in \Cref{tab:ssl_ft}) compared to using \textbf{S} alone, although the gap between them is not statistically significant for metrics other than median (see~\Cref{fig:ssl_fts}).  %

\section{Conclusion}

We present \model, a fully self-supervised (reward-free) approach to representation learning for reinforcement learning, which uses a combination of pretraining objectives to encourage the agent to learn multiple aspects of environment dynamics.
We demonstrate that \model enables significant improvements on the Atari 100k data-efficiency benchmark, especially in comparison to unsupervised exploration approaches which require orders of magnitude more pretraining data.
Investigating the various components of \model, we find that it scales robustly with higher-quality pretraining data and larger models, that all three of the self-supervised objectives contribute to the success of this approach, and that careful reduction of fine-tuning learning rates is critical to optimal performance.

\section{Acknowledgements}
We would like to thank Mila and Compute Canada for computational resources.  We would like to thank Rishabh Agarwal for useful discussions. Aaron Courville would like to acknowledge financial support from Microsoft Research and Hitachi.

\bibliography{main}
\bibliographystyle{icml2021.bst}

\clearpage

\clearpage
\appendix
\section{Reinforcement Learning Background}\label{app:rl_background}
We focus on reinforcement learning in a Markov decision process (MDP).
An agent interacts with an environment, taking actions $a$ and observing states $s$ and rewards $r$.
Interactions are broken up into \textit{episodes}, series of states, actions and rewards that ultimately terminate.
We denote the $t$-th state, action and reward as $s_t$, $a_t$ and $r_t$ respectively, following the convention that $r_t$ is the reward received by the agent after taking action $a_t$ in state $s_t$.
In the control problem examined here, the agent seeks to maximize a discounted sum of rewards $G_t = \sum_{i\ge t} \gamma^i r_i$, where $\gamma \in [0, 1]$ is a discount factor balancing current and future rewards~\citep{Sutton1998}.

\paragraph{Deep Q-Learning}
One common approach to this problem is to estimate $Q(s_t, a) = \mathbf{E}_{\pi^*}[G_t | a_t = a]$.
Assuming that $Q$ is known exactly, the problem of acting optimally is reduced to finding $\max_a Q(s_t, a)$ in each state $s_t$ the agent encounters, which is trivial in environments that possess only a small number of possible actions.
In practice, the true $Q$ can be iteratively approximated by a parameterized $Q_\theta$ with a semi-gradient method, minimizing
\begin{align}
    \mathcal{L}_\theta^{DQN} = (r_t + \gamma \max_a Q_\xi(s_{t+1}, a) - Q_\theta(s_t, a_t))^2
\end{align}
where $Q_\xi$ denotes an older version of $Q_\theta$.
This method has proven extremely successful when used with deep learning, a setting referred to as Deep Q-Networks (DQN)~\citep{dqn}, and more broadly is a common class of deep reinforcement learning (DRL) algorithms.

A number of variants of DQN have been proposed, including those that predict full distributions of future rewards~\citep{bellemare2017distributional}, modifications to the $\max$ operation to reduce value overestimation~\citep{doubledqn}, and architectural modifications to how $Q_\theta$ is predicted ~\citep{wang2016dueling}.
We employ a somewhat modified~\citep{schwarzer2020dataefficient} version of Rainbow~\citep{hessel2018rainbow}, an algorithm that combines many of these innovations.

\section{Uncertainty-aware comparisons}\label{app:uncertainty}
Concurrent work~\citep{rl_precipice} has found that many prior comparisons in deep reinforcement learning are not robust and may be entirely incorrect, particularly in the Atari 100K setting. They demonstrate that these misleading comparisons are partially due to undesirable properties of the per-game median and mean normalized scores, the most commonly-used aggregate metrics, and propose using the inter-quartile mean (IQM) normalized score, calculated over \textit{runs} rather than \textit{tasks}. Moreover, they suggest providing percentile bootstrap confidence intervals to quantify uncertainty, to avoid misleading comparisons based on highly-variable point estimates.

As raw per-run data is required for this, which was not reported for prior work, we do so only for experiments conducted ourselves.  In the interests of improving practices in the community moving forward, we also commit to making this data for our experiments available to other researchers in the future.

\begin{figure}
    \centering
    \begin{subfigure}[b]{0.49\textwidth}
        \centering
        \includegraphics[width=\textwidth]{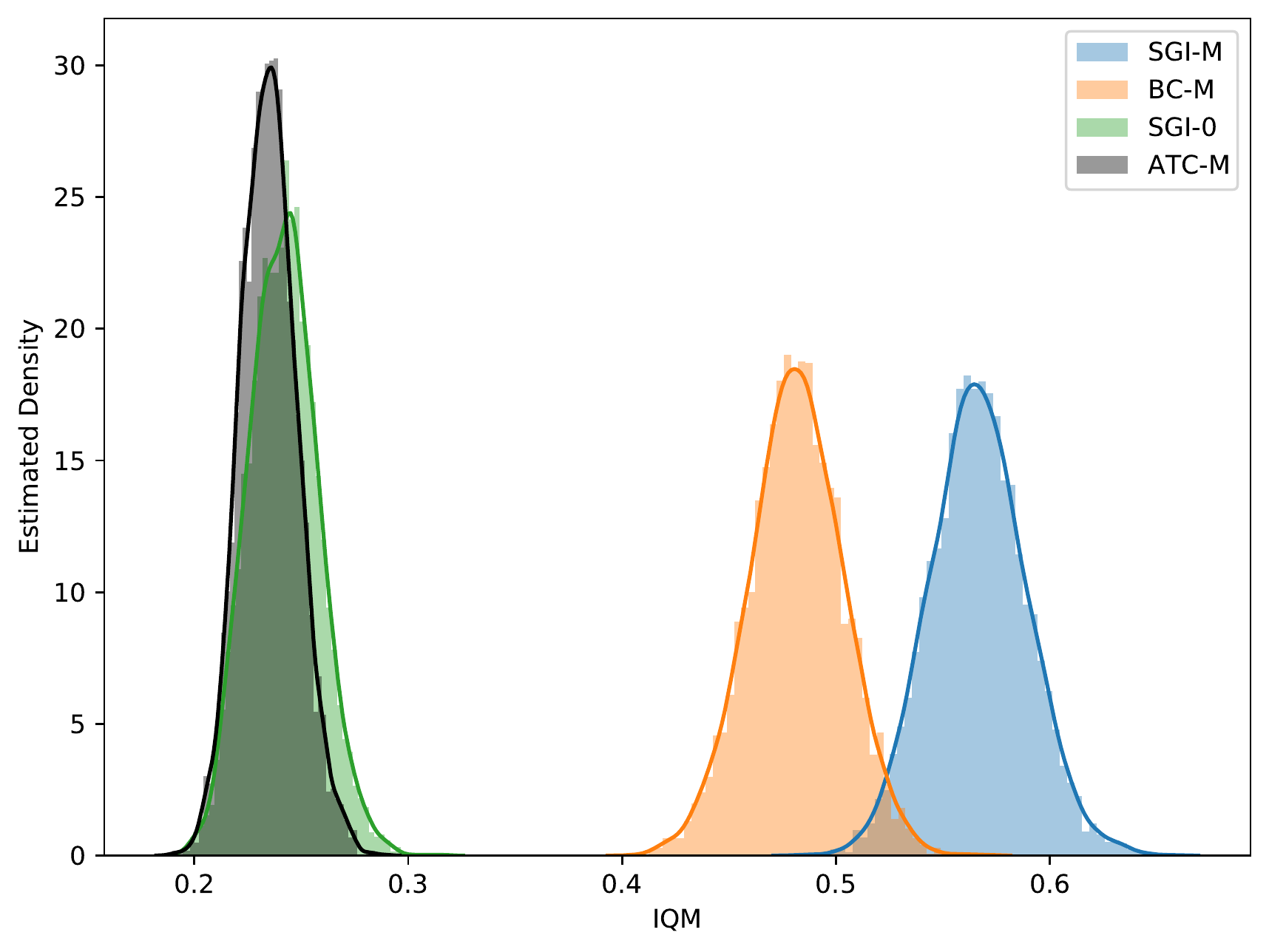}
        \caption{Comparisons to behavioral cloning (BC) and ATC.}
        \label{fig:baseline_comparisons}
    \end{subfigure}
    \begin{subfigure}[b]{0.49\textwidth}
        \centering
        \includegraphics[width=\columnwidth]{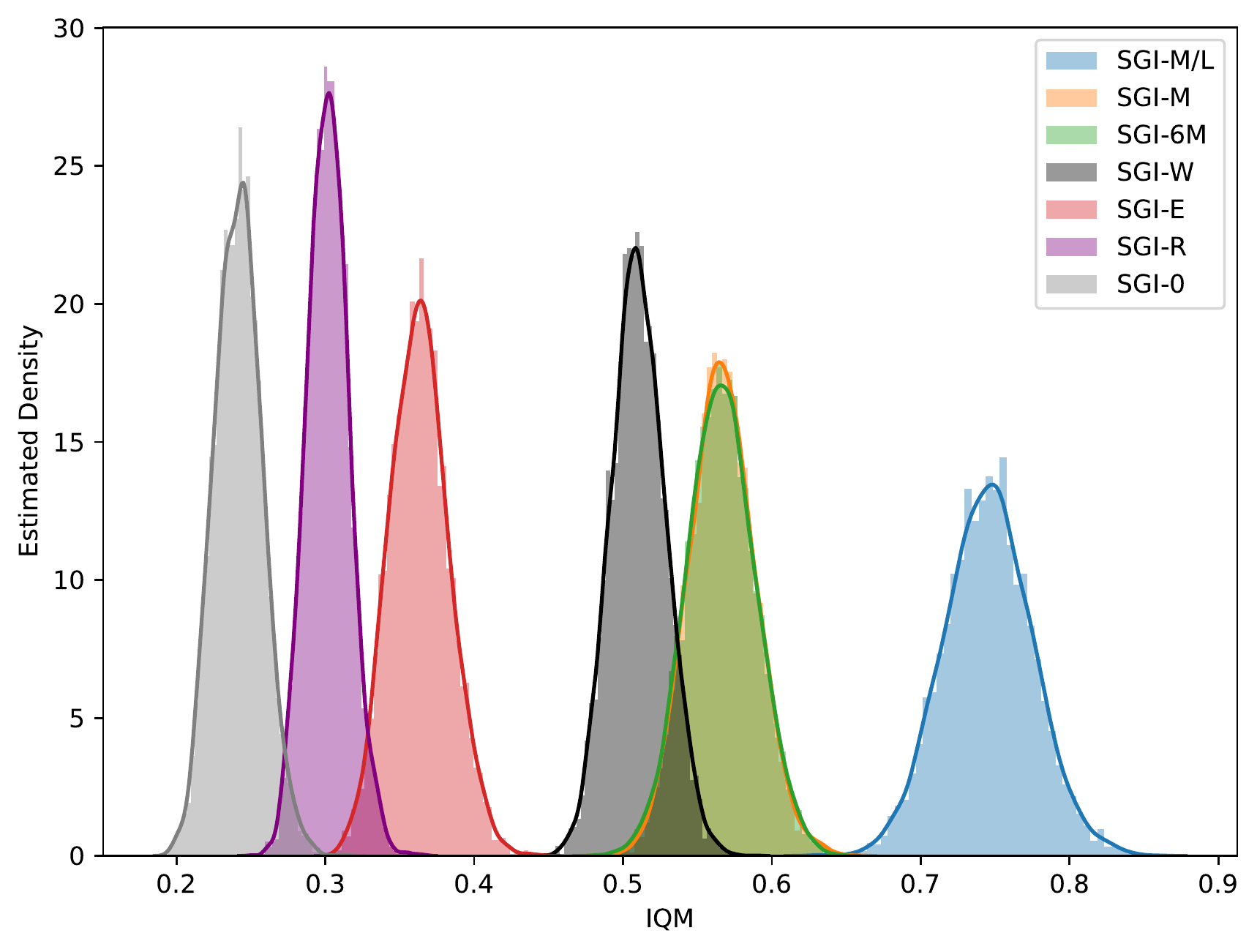}
        \caption{Ablations over different pretraining datasets.}
        \label{fig:sgi_datasets}
    \end{subfigure}
    \begin{subfigure}[b]{0.49\textwidth}
        \includegraphics[width=\columnwidth]{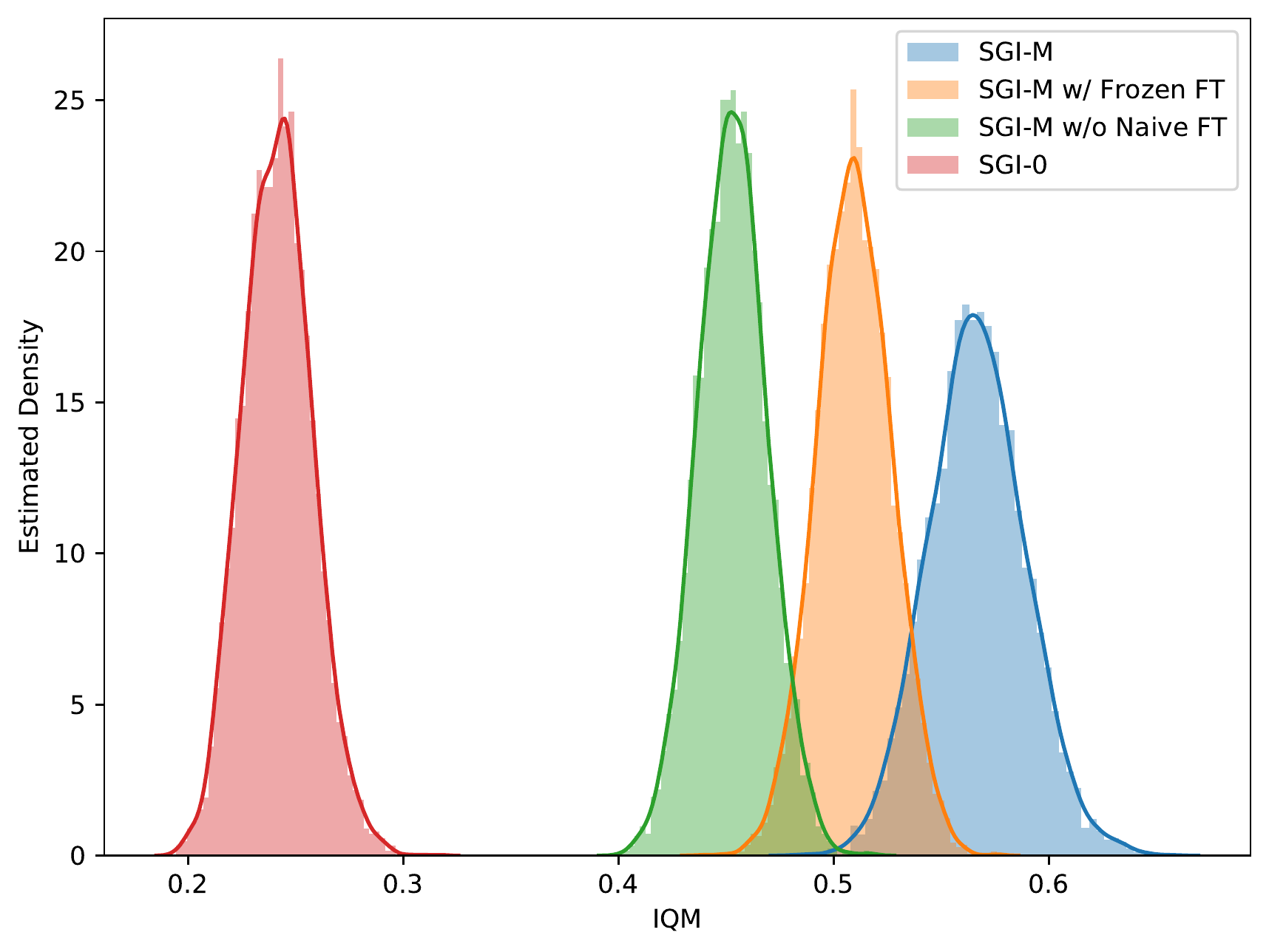}
    \caption{Ablations over various fine-tuning.}
    \label{fig:lr_fts}
    \end{subfigure}
    \begin{subfigure}[b]{0.49\textwidth}
        \centering
        \includegraphics[width=\columnwidth]{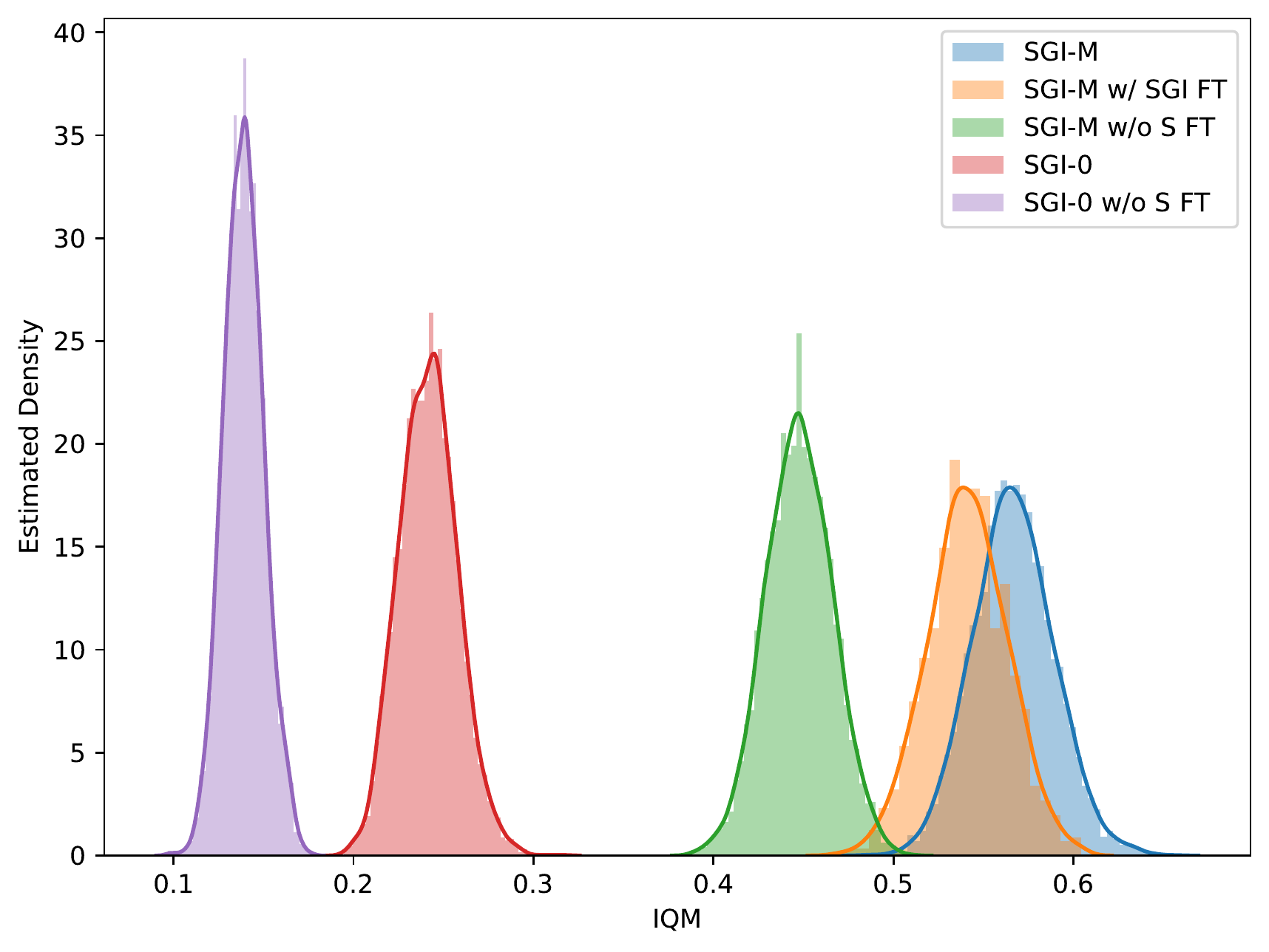}
        \caption{Ablations over SSL objectives during fine-tuning.}
        \label{fig:ssl_fts}
    \end{subfigure}
    \begin{subfigure}[b]{0.49\textwidth}
        \centering
        \includegraphics[width=\columnwidth]{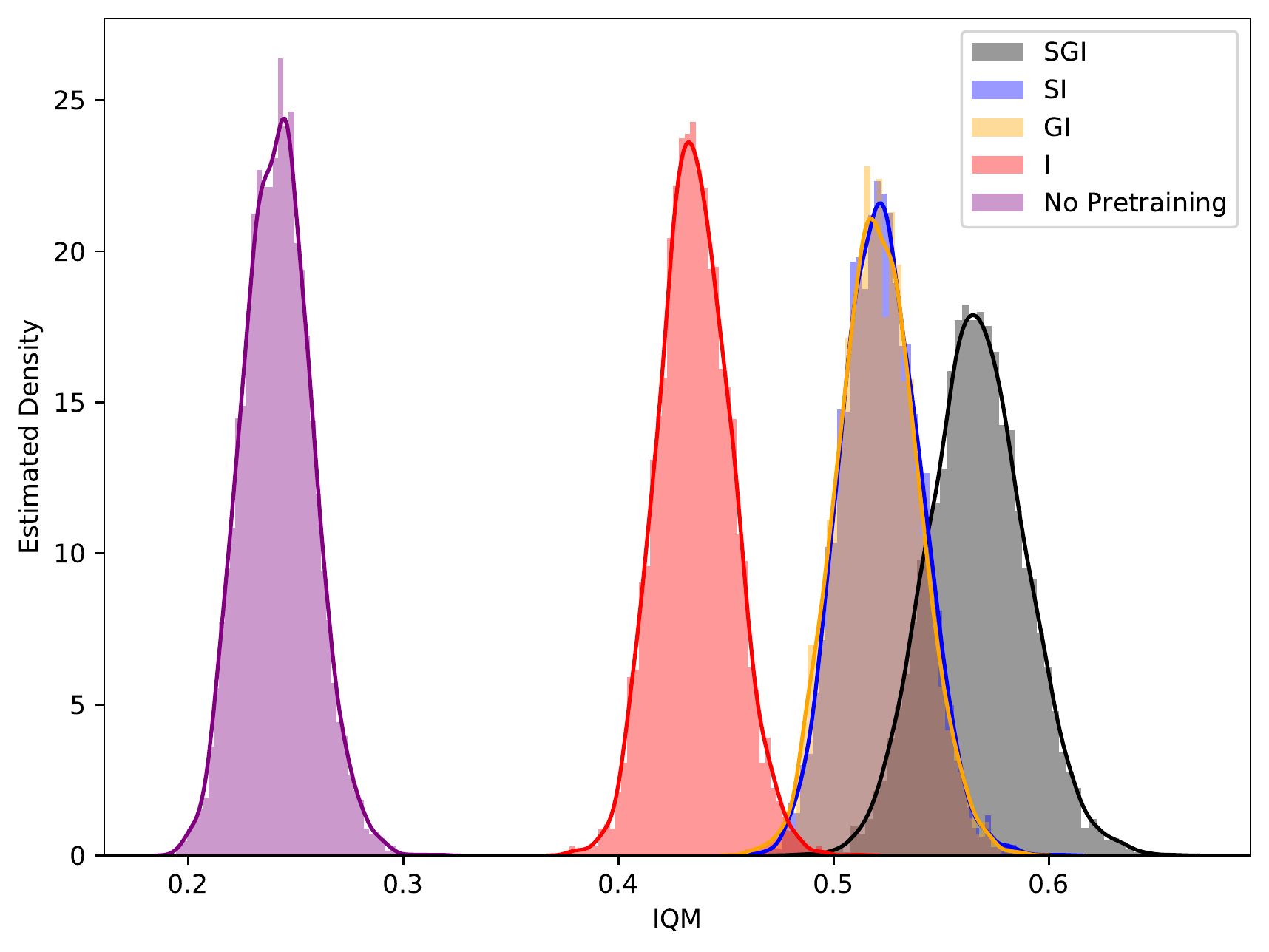}
        \caption{Ablations over pretraining SSL objectives.}
        \label{fig:pretrain_ablations}
    \end{subfigure}
    \caption{Bootstrapping distributions for uncertainty in IQM measurements.}
\end{figure}

In \Cref{fig:baseline_comparisons} through \Cref{fig:pretrain_ablations} we show estimated uncertainty via bootstrapping for the various comparisons drawn throughout~\Cref{sec:discussion}, while Table~\ref{tab:IQMs} gives IQM human-normalized scores and 95\% bootstrap confidence intervals for the same results.  All comparisons in \Cref{fig:baseline_comparisons} through \Cref{fig:pretrain_ablations} are statistically significant ($p<0.05$) except for:
\begin{itemize}
    \item ATC-M vs SGI-None in \Cref{fig:baseline_comparisons} ($p \gg 0.05$)
    \item SGI-M vs SGI-W in \Cref{fig:sgi_datasets} ($p \approx 0.05$)
    \item SGI-M vs SGI-M w/ SGI FT in \Cref{fig:ssl_fts} ($p \approx 0.4$)
    \item SGI-M vs G+I and S+I in \Cref{fig:pretrain_ablations} ($p \approx 0.1$)
\end{itemize}

\begin{table}[]
    \centering
\caption{Interquartile mean, median and mean human-normalized scores for variants of SGI and controls, evaluated after finetuning over all 10 runs for each of the 26 Atari 100k games.  Confidence intervals computed by percentile bootstrap with 5000 resamples.}
\label{tab:IQMs}
    \begin{tabular}{lrrrrrr}
    \toprule
    Method & IQM & 95\% CI & Median & 95\% CI & Mean & 95\% CI \\
    \midrule
SGI-M/L  & \textbf{0.745} & (0.687, 0.805) & \textbf{0.753} & (0.625, 0.850) & \textbf{1.598} & (1.486, 1.676)\\
SGI-M  & 0.567 & (0.524, 0.612) &0.679 & (0.473, 0.739) & 1.149 & (0.974, 1.347)\\
SGI-M/S  & 0.444 & (0.404, 0.487) & 0.423 & (0.341, 0.577) & 0.914 & (0.822, 1.031)\\
SGI-W  & 0.510 & (0.476, 0.547) &0.589 & (0.434, 0.675) & 1.144 & (0.981, 1.345)\\
SGI-E  & 0.363 & (0.326, 0.404) &0.456 & (0.309, 0.482) & 0.838 & (0.692, 1.008)\\
SGI-R  & 0.302 & (0.275, 0.331) &0.326 & (0.253, 0.385) & 0.888 & (0.776, 1.004)\\
SGI-None  & 0.242 & (0.212, 0.274) &0.343 & (0.268, 0.401) & 0.565 & (0.440, 0.711)\\
\midrule
\multicolumn{3}{l}{ \textit{Baselines}} \\
\midrule
ATC-M  & 0.235 & (0.210, 0.262) & 0.204 & (0.182, 0.291) & 0.780 & (0.601, 0.971)\\
ATC-W  & 0.221 & (0.199, 0.244)   & 0.219 & (0.170, 0.290) & 0.587 & (0.504, 0.673)\\
ATC-E  & 0.214 & (0.193, 0.236)   & 0.237 & (0.169, 0.266) & 0.462 & (0.420, 0.504)\\
ATC-R  & 0.187 & (0.174, 0.202)   & 0.191 & (0.139, 0.202) & 0.472 & (0.454, 0.491)\\
BC-M  & 0.481 & (0.438, 0.524)   &0.548 & (0.390, 0.685) & 0.858 & (0.795, 0.924)\\
\midrule
\multicolumn{3}{l}{ \textit{Pretraining Ablations}} \\
\midrule
S+I  & 0.522 & (0.488, 0.559) & 0.629 & (0.494, 0.664)& 0.978 & (0.900, 1.061)\\
G+I  & 0.521 & (0.486, 0.558) & 0.512 & (0.386, 0.582)& 1.004 & (0.892, 1.129)\\
S+G  & 0.032 & (0.027, 0.039) & 0.029 & (0.025, 0.044)& 0.098 & (0.061, 0.146)\\
I  & 0.435 & (0.404, 0.470) &0.411 & (0.334, 0.489)& 0.943 & (0.783, 1.126)\\
G  & 0.060 & (0.048, 0.072) &0.060 & (0.037, 0.081)& 0.181 & (0.145, 0.218)\\
S  & 0.007 & (0.002, 0.011) &0.009 & (0.002, 0.014)& -0.054 & (-0.082, -0.026)\\
\midrule
\multicolumn{3}{l}{ \textit{Finetuning Ablations}} \\
\midrule
SGI-M (No S) & 0.448 & (0.412, 0.484)  &  0.419 & (0.335, 0.524)& 1.114 & (0.921, 1.321)\\
SGI-None (No S) & 0.139 & (0.118, 0.162) & 0.161 & (0.123, 0.225) & 0.315 & (0.274, 0.356)\\
SGI-M (All SGI)  & 0.541 & (0.498, 0.585)  &  0.397 & (0.330, 0.503)&1.011 & (0.909, 1.071)\\
SGI-M (Frozen)  & 0.510 & (0.476, 0.543)  &  0.499 & (0.406, 0.554)&  0.971 & (0.871, 1.088)\\
SGI-M (Naive)  & 0.453 & (0.422, 0.485) & 0.429 & (0.380, 0.500)& 0.845 & (0.754, 0.952)\\
\bottomrule
\end{tabular}
\end{table}

\section{Implementation Details}\label{app:implementation}

We base our work on the code released for SPR \citep{schwarzer2020dataefficient}, which in turn is based on rlpyt~\citep{rlpyt}, and makes use of NumPy~\citep{harris2020array} and PyTorch~\citep{pytorch}.

\subsection{Training}
We set $\lambda^\text{SPR}=2$ and $\lambda^\text{IM}=1$ during pre-training.
Unless otherwise noted, all settings match SPR during fine-tuning, including batch size, replay ratio, target network update period, and $\lambda^\text{SPR}$.
We use a batch size of 256 during pre-training to maximize throughput, and update both the SPR and goal-conditioned RL target network target networks with an exponential moving average with $\tau = 0.99$.
We pre-train for a number of gradient steps equivalent to 10 epochs over 6M samples, no matter the amount of data used.
Due to the cost of pretraining, we pre-train a single encoder per game for each configuration tested.
However, we use 10 random seeds at fine-tuning time, allowing us to average over variance due to exploration and data order.
Finally, we reduce fine-tuning learning rates for pretrained encoders and dynamics models by a factor of 100, and by a factor of 3 for other pretrained weights.
We find this crucial to \model's performance, and discuss it in detail in Section~\ref{sec:learning_rates}.

We trained \model on standard GPUs, including V100s and P100s.  We found that pretraining took roughly one to three days and finetuning between four and 12 hours per run on a single GPU, depending on the size of the network used and type of GPU.

\subsection{Goal-Conditioned Reinforcement Learning}\label{app:gcrl}
We generate goals in a three-stage process: a goal $g$ for state $s_t$ is initially chosen to be the target representation of a state sampled uniformly from the near future, $g \gets \tilde{z}_{t+i}, i \sim\text{Uniform}(50)$, before being combined with a normalized vector of isotropic Gaussian noise $n$ as $g \gets \alpha n + (1 - \alpha) g$, where $\alpha \sim \text{Uniform}(0, 0.5)$.
Finally, we exchange goal vectors between states in the minibatch with probability 0.2, to ensure that some goals correspond to states reached in entirely different trajectories.

In defining our synthetic goal-conditioned rewards, we take inspiration from potential-based reward shaping~\citep{ng1999policy}.
Using the target representations $\tilde{z}_t \triangleq f_m(s_t)$ and $\tilde{z}_{t+1} \triangleq f_m(s_{t+1})$, we define the reward as follows: \begin{flalign}
    R(\tilde{z}_{t},\tilde{z}_{t+1}, g) = d(\tilde{z}_{t}, g) - d(\tilde{z}_{t+1}, g) \\
    d(\tilde{z}_t, g) = \exp{\left (2\frac{\tilde{z}_{t} \cdot g}{||\tilde{z}||_2 \cdot  || g||_2}  - 2 \right ).}
\end{flalign}
As this reward function depends on the target encoder $f_m$, it changes throughout training, although using the slower-moving $f_m$ rather than the online encoder $f_o$ may provide some measure of stability.
Like SPR, however, this objective is technically vulnerable to collapse.
If all representations $\tilde{z}_t$ collapse to a single constant vector then all rewards will be 0, allowing the task to be trivially solved.

We estimate $Q(s_t, a_t, g)$ using FiLM~\citep{perez2018film} to condition the DQN on the goal $g$, which we found to be more robust than simple concatenation.
A FiLM generator $j$ produces per-channel biases $\beta_c$ and scales $\gamma_c$, which then modulate features through a per-channel affine transformation:
\begin{align}
    \text{FiLM}(F_c|\gamma_c, \beta_c) = \gamma_c F_c + \beta_c
\end{align}
We use these parameters to replace the learned per-channel affine transformation in a layer norm layer \citep{ba2016layer}, which we insert immediately prior to the final linear layer in the DQN head.

We apply FiLM after the first layer in the DQN's MLP head.
We parameterize our FiLM generator $j$ as a small convolutional network, which takes the goal $g$ (viewed as a $64 \times 7 \times 7$ spatial feature map) as input and applies two 128-channel convolutions followed by a flatten and linear layer to produce the FiLM parameters $\gamma$ and $\beta$. 

\subsection{Model Architectures}
In addition to the standard three-layer CNN encoder introduced by \cite{dqn}, we experiment with larger residual networks~\citep{resnet}.  We use the design proposed by~\citet{impala} as a starting point, while still adopting innovations used in more modern architectures such as EfficientNets~\citep{efficientnet} and MobileNetv2~\citep{sandler2018mobilenetv2}.
In particular, we use inverted residual blocks with an expansion ratio of 2, and batch normalization~\citep{ioffe2015batch} after each convolutional layer.
We use three groups of three residual blocks with 32, 64 and 64 channels each, downscaling by a factor of three in the first group and two in each successive group.  This yields a final representation of shape $64 \times 7 \times 7$ when applied to $84 \times 84$-dimensional Atari frames, identical to that of the standard CNN encoder.  In our scaling experiment with a larger network, we increase to five blocks per group, with 48, 96 and 96 channels in each group, as well as using a larger expansion ratio of 4, producing a representation of shape $96 \times 7 \times 7$.
This enlargement increases the number of parameters by roughly a factor of 5.   Finally, our DQN head has 512 hidden units, as opposed to 256 in SPR.

\subsection{Image Augmentation}
We use the same image augmentations as used in SPR \citep{schwarzer2020dataefficient}, which itself used the augmentations used in DrQ~\citep{drq}, in all experiments, including during both pretraining and fine-tuning.
Specifically, we employ random crops (4 pixel padding and 84x84 crops) in combination with image intensity jittering.

\subsection{Experiments with ATC}\label{app:atc}
As ATC~\citep{stooke2020decoupling} was not tested on the Atari100k setting, and as its hyperparameters (including network size and fine-tuning scheme) are very different from those used by \model, we modify its code\footnote{ \url{https://github.com/astooke/rlpyt/tree/master/rlpyt/ul}} to allow it to be fairly compared to \model.  We replace the convolutional encoder with that used by \model, and use the same optimizer settings, image augmentation, pre-training data, and number of pre-training epochs as in \model.  However, we retain ATC's mini-batch structure (i.e., sampling 32 subsequences of eight consecutive time steps, for a total batch size of 512), as this structure defines the negative samples used by ATC's InfoNCE loss.  During fine-tuning, we transfer the ATC projection head to the first layer of the DQN MLP head, as in SPR; we otherwise fine-tune identically to \model, including using SPR.

\section{Pseudocode}\label{pseudocode}
\begin{minipage}{\textwidth}
\begin{algorithm}[H]
 Denote parameters of online encoder $f_o$, projection $p_o$ and Q-learning head as $\theta_o$\;
 
 Denote parameters of target encoder $f_m$, projection $p_m$ and Q-learning target head as $\theta_m$\;
 
 Denote parameters of transition model $h$, predictor $q$, inverse model $I$ as $\phi$\;
 
 Denote the maximum prediction depth as $K$, batch size as $N$\;
 
 Denote distance function in goal RL reward as $d$\;
 
 initialize offline dataset $D$\;
 
 \While{Training}{
 sample a minibatch of sequences of $(s_t, a, s+{t+1}) \sim D$ \tcp*{sample unlabeled data}
 
 \tcc{sample goals}
 \For{$i$ in range$(0, N)$}{
  $s^i \gets \text{augment}(s^i); {s'}^i \gets \text{augment}({s'}^i)$ \tcp*{augment input images}
  
  $j \sim \text{Discrete Uniform}(1, 50)$ \tcp*{sample hindsight goal states}
  $g^i \gets f_m(s^n_j)$ \tcp*{encode goal states}
  $\alpha \sim \text{Uniform}(0, 0.5)$,
  $n \sim \text{Normal}(0, 1)$ \tcp*{sample noise parameters}
  $g^i \gets \alpha g^i + (1 - \alpha)n$ \tcp*{apply noise}
  \tcc{Permute to make some goals very challenging to reach}
  permute$ ~\sim \text{Bernoulli}(0.2)$
  
  \If{permute}{
    $j \sim \text{Discrete Uniform}(N)$
    
    $g^i \gets g^j$ \tcp*{permute goal}
  }}

\tcc{compute \model loss} 
 \For{$i$ in range$(0, N)$}{
  
  $\hat{z}^i_0 \gets f_\theta(s^i_0)$ \tcp*{compute online representations}
  $l^i \gets 0$\;

  \tcc{compute SPR loss}
  \For{k in (1, \ldots, K)}{
    $\hat{z}^i_k \gets h(\hat{z}^i_{k-1}, a^i_{k-1})$ \tcp*{latent states via transition model}
    $\tilde{z}^i_k \gets f_{m}(s^i_k)$ \tcp*{target representations}
    $\hat{y}^i_k \gets q(p_o(\hat{z}^i_k))$,
    $\tilde{y}^i_k \gets g_m(\tilde{z}^i_k)$ \tcp*{projections}
    $l^i \gets l^i - \lambda^{\text{SPR}}\left ( \frac{\tilde{y}^i_{k}}{||\tilde{y}^i_{k}||_2} \right)^{\top}\left ( \frac{{\hat{y}^i}_{k}}{||{\hat{y}^i}_{k}||_2} \right )$ \tcp*{\model loss at step  $k$}
  }
  \tcc{compute inverse modeling loss}
  \For{k in (1, \ldots, K)}{
    $l^i \gets \lambda^{\text{IM}} \cdot \text{Cross-entropy loss} (a^i_{k-1}, I(\hat{y}_{k-1}, \tilde{y}_{k}))$
  }
  
  \tcc{compute goal RL loss}
  $r^i \gets d(g^i, \tilde{z}_t) - d(g^i, \tilde{z}_{t+1})$ \tcp*{Calculate goal RL reward}
  $l^i \gets l^i + \text{RL loss}(s^i, a^i, r^i, {s'}^i)$ \tcp*{Add goal RL loss for batch}%
  }
  $l \gets \frac{1}{N}\sum_{i=0}^{N} l^i$ \tcp*{average loss over minibatch}
  $\theta_o, \phi \gets \text{optimize}((\theta_o, \phi), l)$ \tcp*{update online parameters}
  $\theta_m \gets \tau\theta_o + (1 - \tau)\theta_m$ \tcp*{update target parameters}
}
return $(\theta_o,\; \phi)$ \tcp*{return parameters for fine-tuning}
\caption{Pre-Training with \model}
\end{algorithm}
\end{minipage}

\FloatBarrier
\newpage
\clearpage
\newpage

\section{Full Results on Atari100k}\label{app:full_results}

We report full scores for \model agents across all 26 games in \Cref{tab:all_scores}.  We do not reproduce the per-game scores for APT and VISR provided by \citet{liu2021unsupervised}, as we believe that the scores in the currently-available version of their paper may contain errors.\footnote{In particular, we observed that VISR claimed to have a score below $-21$ on Pong, which is impossible with standard settings.} 
\begin{table*}[h]
\caption{
Mean return per episode for the 26 Atari100k games~\citep{simple} after 100k steps.
Agents are evaluated at the end of training, and scores for all methods are averaged over 10 random seeds.
We reproduce scores for SPR from \citet{schwarzer2020dataefficient}, whereas ATC scores are from our implementation.
}
\label{tab:all_scores}
\centering
\scalebox{0.8}{
\footnotesize{
\begin{tabular}{lllllllllll}
\toprule
{} &   Random &    Human &      SPR &      ATC-M &    SGI-R &    SGI-E &    SGI-W &  SGI-M/S &    SGI-M &  SGI-M/L \\
\midrule
Alien           &    227.8 &   7127.7 &    801.5 &    699.0 &   1034.5 &    857.6 &   1043.8 &   1070.5 &   1101.7 &\textbf{   1184.0 }  \\
Amidar          &      5.8 &   1719.5 &    176.3 &     95.4 &    154.8 &    166.8 &\textbf{    206.7 }&    185.9 &    168.2 &    171.2   \\
Assault         &    222.4 &    742.0 &    571.0 &    509.8 &    446.6 &    583.1 &    759.5 &    632.4 &    905.1 &\textbf{   1326.5 }  \\
Asterix         &    210.0 &   8503.3 &    977.8 &    454.1 &    754.6 &    953.6 &\textbf{   1539.1 }&    651.8 &    835.6 &    567.2   \\
Bank Heist      &     14.2 &    753.1 &    380.9 &    534.9 &    397.4 &    514.8 &    426.3 &    547.4 &\textbf{    608.4 }&    567.8   \\
Battle Zone     &   2360.0 &  37187.5 &\textbf{  16651.0 }&  13683.8 &   4439.0 &  16417.0 &   7103.0 &  12107.0 &  13170.0 &  14462.0   \\
Boxing          &      0.1 &     12.1 &     35.8 &     16.8 &     57.7 &     33.6 &     50.2 &     40.0 &     36.9 &\textbf{     73.9 }  \\
Breakout        &      1.7 &     30.5 &     17.1 &     16.9 &     23.4 &     17.8 &     35.4 &     23.8 &     42.8 &\textbf{    251.9 }  \\
Chopper Command &    811.0 &   7387.8 &    974.8 &    870.8 &    784.7 &   1136.2 &   1040.1 &   1042.7 &\textbf{   1404.0 }&   1037.9   \\
Crazy Climber   &  10780.5 &  35829.4 &  42923.6 &  74215.5 &  50561.2 &  76356.3 &  81057.4 &  75542.1 &  88561.2 &\textbf{  94602.2 }  \\
Demon Attack    &    152.1 &   1971.0 &    545.2 &    524.6 &   2198.7 &    357.5 &   1408.5 &   1135.5 &    968.1 &\textbf{   5634.8 }  \\
Freeway         &      0.0 &     29.6 &     24.4 &      5.7 &      2.1 &     15.1 &     26.5 &     12.5 &\textbf{     30.0 }&     28.6   \\
Frostbite       &     65.2 &   4334.7 &\textbf{   1821.5 }&    222.6 &    349.3 &    981.4 &    247.7 &    861.1 &    741.3 &    927.8   \\
Gopher          &    257.6 &   2412.5 &    715.2 &    946.2 &   1033.9 &    964.9 &   1846.0 &   1172.4 &   1660.4 &\textbf{   2035.8 }  \\
Hero            &   1027.0 &  30826.4 &   7019.2 &   6119.4 &   7875.2 &   6863.7 &   7503.9 &   7090.4 &   7474.0 &\textbf{   9975.9 }  \\
Jamesbond       &     29.0 &    302.8 &    365.4 &    272.6 &    263.9 &    383.8 &\textbf{    425.1 }&    413.2 &    366.4 &    394.8   \\
Kangaroo        &     52.0 &   3035.0 &\textbf{   3276.4 }&    603.1 &    923.8 &   1588.9 &    598.6 &   1236.8 &   2172.8 &   1887.5   \\
Krull           &   1598.0 &   2665.5 &   3688.9 &   4494.7 &   5672.6 &   4070.7 &   5583.2 &\textbf{   6161.3 }&   5734.0 &   5862.6   \\
Kung Fu Master  &    258.5 &  22736.3 &  13192.7 &  11648.2 &  13349.2 &  11802.1 &  14199.7 &  16781.8 &  16137.8 &\textbf{  17340.7 }  \\
Ms Pacman       &    307.3 &   6951.6 &   1313.2 &    848.9 &    411.0 &   1278.3 &   1970.8 &   1519.5 &   1520.0 &\textbf{   2218.0 }  \\
Pong            &    -20.7 &     14.6 &     -5.9 &    -13.5 &     -3.9 &      4.2 &      4.7 &\textbf{      9.7 }&      7.6 &      7.7   \\
Private Eye     &     24.9 &  69571.3 &\textbf{    124.0 }&     95.0 &     95.3 &    100.0 &    100.0 &     84.7 &     90.0 &     83.8   \\
Qbert           &    163.9 &  13455.0 &    669.1 &    572.2 &    595.0 &    717.6 &\textbf{    855.6 }&    804.7 &    709.8 &    702.6   \\
Road Runner     &     11.5 &   7845.0 &  14220.5 &   7989.3 &   5476.0 &   9195.2 &  18011.9 &  12083.5 &\textbf{  18370.2 }&  18306.8   \\
Seaquest        &     68.4 &  42054.7 &    583.1 &    415.7 &    735.3 &    615.2 &    656.1 &    728.2 &    728.4 &\textbf{   1979.3 }  \\
Up N Down       &    533.4 &  11693.2 &  28138.5 &  84361.2 &  67968.1 &  63612.9 &\textbf{  84551.4 }&  42165.6 &  79228.8 &  46083.3   \\ \midrule
Median HNS      &    0.000 &    1.000 &    0.415 &    0.204 &    0.326 &    0.456 &    0.589 &    0.423 &    0.679 &\textbf{    0.755 }  \\
Mean HNS        &    0.000 &    1.000 &    0.704 &    0.780 &    0.888 &    0.838 &    1.144 &    0.914 &    1.149 &\textbf{    1.590 }  \\ \midrule
\#Games $>$ Human  &        0 &        0 &        7 &        5 &        5 &        6 &        8 &        6 &\textbf{        9 }&\textbf{        9 }  \\
\#Games $>$ 0      &        0 &       26 &\textbf{       26 }&\textbf{       26 }&       25 &\textbf{       26 }&\textbf{       26 }&\textbf{       26 }&\textbf{       26 }&\textbf{       26 }  \\

\bottomrule
\end{tabular}
}
}
\end{table*}

\begin{table*}[ht]
    \caption{Mean return per episode for the 26 Atari100k games~\citep{simple} after 100k steps for versions of \model with modified fine-tuning, as discussed in \Cref{sec:discussion}. Agents are evaluated at the end of training, and scores for all methods are averaged over 10 random seeds.  We reproduce scores for SPR from \citet{schwarzer2020dataefficient}.}
    \label{tab:all_ft_control_scores}
    \centering
    \scalebox{0.8}{\begin{tabular}{lllllllll}
\toprule
{} &   Random &    Human & SGI-None &    Naive &   Frozen &    No SPR & Full SSL &    SGI-M \\
\midrule
Alien           &    227.8 &   7127.7 &    835.9 &   1049.3 &\textbf{   1242.8 }&    1060.7 &   1117.6 &   1101.7   \\
Amidar          &      5.8 &   1719.5 &    107.6 &    133.6 &    147.7 &     154.2 &\textbf{    206.0 }&    168.2   \\
Assault         &    222.4 &    742.0 &    657.7 &    752.1 &    869.2 &     756.3 &\textbf{   1145.2 }&    905.1   \\
Asterix         &    210.0 &   8503.3 &    832.9 &\textbf{   1029.3 }&    433.1 &     575.5 &    603.1 &    835.6   \\
Bank Heist      &     14.2 &    753.1 &    613.2 &\textbf{    726.5 }&    273.6 &     365.8 &    323.4 &    608.4   \\
Battle Zone     &   2360.0 &  37187.5 &  13490.0 &\textbf{  15708.0 }&  11754.0 &   13692.0 &  11689.8 &  13170.0   \\
Boxing          &      0.1 &     12.1 &      6.6 &     24.0 &\textbf{     61.5 }&      34.7 &     42.7 &     36.9   \\
Breakout        &      1.7 &     30.5 &     12.1 &     29.3 &     34.0 &      43.0 &\textbf{     62.6 }&     42.8   \\
Chopper Command &    811.0 &   7387.8 &   1085.2 &   1081.2 &    916.5 &     925.5 &    965.8 &\textbf{   1404.0 }  \\
Crazy Climber   &  10780.5 &  35829.4 &  19707.6 &  55002.4 &  65220.0 &   69505.6 &  69052.0 &\textbf{  88561.2 }  \\
Demon Attack    &    152.1 &   1971.0 &    778.8 &    850.0 &   1329.4 &     981.7 &\textbf{   1783.8 }&    968.1   \\
Freeway         &      0.0 &     29.6 &     17.2 &     28.1 &     24.4 &      13.2 &     10.9 &\textbf{     30.0 }  \\
Frostbite       &     65.2 &   4334.7 &   1475.8 &    662.1 &   1045.4 &     482.1 &\textbf{   1664.9 }&    741.3   \\
Gopher          &    257.6 &   2412.5 &    438.2 &    626.1 &\textbf{   2214.1 }&    1561.7 &   1998.7 &   1660.4   \\
Hero            &   1027.0 &  30826.4 &   6472.0 &   5538.3 &   6353.3 &    5249.6 &\textbf{   8715.4 }&   7474.0   \\
Jamesbond       &     29.0 &    302.8 &    157.4 &    324.2 &    358.2 &     346.8 &\textbf{    407.6 }&    366.4   \\
Kangaroo        &     52.0 &   3035.0 &\textbf{   3802.8 }&   3091.6 &    800.0 &     685.6 &    999.5 &   2172.8   \\
Krull           &   1598.0 &   2665.5 &   3954.0 &   5202.7 &\textbf{   6073.7 }&    5722.8 &   5323.9 &   5734.0   \\
Kung Fu Master  &    258.5 &  22736.3 &   7929.4 &  11952.2 &\textbf{  19374.6 }&   15039.8 &  18123.2 &  16137.8   \\
Ms Pacman       &    307.3 &   6951.6 &    990.2 &   1276.4 &   1663.3 &    1753.3 &\textbf{   1779.3 }&   1520.0   \\
Pong            &    -20.7 &     14.6 &     -4.4 &     -4.2 &      3.8 &       3.9 &     -0.1 &\textbf{      7.6 }  \\
Private Eye     &     24.9 &  69571.3 &     62.8 &\textbf{    385.9 }&     96.7 &      90.5 &     90.0 &     90.0   \\
Qbert           &    163.9 &  13455.0 &    720.0 &    664.8 &    587.6 &     681.3 &\textbf{   3015.8 }&    709.8   \\
Road Runner     &     11.5 &   7845.0 &   5428.4 &  14629.7 &  14311.9 &   17036.5 &  13998.2 &\textbf{  18370.2 }  \\
Seaquest        &     68.4 &  42054.7 &    577.8 &    509.0 &   1054.4 &\textbf{    1397.8 }&    989.4 &    728.4   \\
Up N Down       &    533.4 &  11693.2 &  46042.6 &  48856.6 &  29938.4 &\textbf{  105466.9 }&  45023.5 &  79228.8   \\ \midrule
Median HNS      &    0.000 &    1.000 &    0.343 &    0.425 &    0.499 &     0.452 &    0.397 &\textbf{    0.679 }  \\
Mean HNS        &    0.000 &    1.000 &    0.565 &    0.849 &    0.971 &     1.114 &    1.011 &\textbf{    1.149 }  \\ \midrule
\#Games $>$ Human  &        0 &        0 &        3 &        8 &        8 &         8 &        8 &\textbf{        9 }  \\
\#Games $>$ SPR    &        0 &       19 &       10 &       14 &       15 &        14 &       17 &\textbf{       20 }  \\
\bottomrule
\end{tabular}
}
\end{table*}

\begin{table}[ht]
\caption{Mean return per episode for the 26 Atari100k games~\citep{simple} after 100k steps for various combinations of \model's pretraining objectives, as discussed in \Cref{sec:discussion}.  Agents are evaluated at the end of training, and scores for all methods are averaged over 10 random seeds.}\label{tab:all_abl_scores}
\centering
    \scalebox{0.735}{
\begin{tabular}{lllllllllll}
\toprule
{} &   Random &    Human &     None &       S &        G &        I &     G+I &      S+G &      S+I &      SGI \\
\midrule
Alien           &    227.8 &   7127.7 &    835.9 &   278.7 &    964.3 &   1161.6 &   571.2 &   1172.3 &\textbf{   1203.0 }&   1101.7   \\
Amidar          &      5.8 &   1719.5 &    107.6 &    37.8 &     54.8 &    198.1 &    58.0 &\textbf{    210.5 }&    175.4 &    168.2   \\
Assault         &    222.4 &    742.0 &    657.7 &   517.9 &    512.3 &    868.1 &   567.2 &    813.5 &    820.3 &\textbf{    905.1 }  \\
Asterix         &    210.0 &   8503.3 &    832.9 &   292.6 &    416.1 &    475.6 &   431.8 &    506.3 &    648.5 &\textbf{    835.6 }  \\
Bank Heist      &     14.2 &    753.1 &\textbf{    613.2 }&     3.1 &    115.2 &    357.6 &    57.2 &    423.3 &    547.5 &    608.4   \\
Battle Zone     &   2360.0 &  37187.5 &  13490.0 &  4665.0 &   3336.0 &  14807.0 &  3249.0 &  12528.0 &\textbf{  15491.0 }&  13170.0   \\
Boxing          &      0.1 &     12.1 &      6.6 &   -21.8 &     12.5 &     40.1 &    -0.4 &\textbf{     42.9 }&     38.3 &     36.9   \\
Breakout        &      1.7 &     30.5 &     12.1 &     0.9 &      2.1 &     24.1 &     3.2 &     41.0 &     41.6 &\textbf{     42.8 }  \\
Chopper Command &    811.0 &   7387.8 &   1085.2 &   799.7 &    813.1 &    973.1 &   923.7 &   1097.2 &    978.3 &\textbf{   1404.0 }  \\
Crazy Climber   &  10780.5 &  35829.4 &  19707.6 &   243.3 &  17760.3 &  51203.9 &   581.0 &  66228.5 &  83995.4 &\textbf{  88561.2 }  \\
Demon Attack    &    152.1 &   1971.0 &    778.8 &   668.9 &    316.9 &\textbf{   1524.6 }&   756.4 &   1008.4 &   1286.6 &    968.1   \\
Freeway         &      0.0 &     29.6 &     17.2 &    15.2 &     17.7 &      2.6 &    19.3 &\textbf{     30.5 }&     29.1 &     30.0   \\
Frostbite       &     65.2 &   4334.7 &\textbf{   1475.8 }&   427.2 &    523.3 &    395.0 &   215.4 &    530.5 &    463.3 &    741.3   \\
Gopher          &    257.6 &   2412.5 &    438.2 &    60.7 &    129.0 &\textbf{   1966.1 }&    99.0 &   1747.4 &   1778.7 &   1660.4   \\
Hero            &   1027.0 &  30826.4 &   6472.0 &  2381.2 &   3590.2 &   7177.6 &  3998.7 &\textbf{   8251.2 }&   7366.2 &   7474.0   \\
Jamesbond       &     29.0 &    302.8 &    157.4 &    41.8 &    236.0 &    373.1 &   183.6 &    365.6 &\textbf{    378.4 }&    366.4   \\
Kangaroo        &     52.0 &   3035.0 &\textbf{   3802.8 }&   129.8 &    401.6 &   1041.4 &   222.6 &    830.8 &    760.2 &   2172.8   \\
Krull           &   1598.0 &   2665.5 &   3954.0 &   720.1 &   1241.4 &\textbf{   5859.8 }&  1582.4 &   5778.8 &   5808.6 &   5734.0   \\
Kung Fu Master  &    258.5 &  22736.3 &   7929.4 &    79.7 &    453.7 &  16914.7 &   686.2 &\textbf{  17825.1 }&  14681.9 &  16137.8   \\
Ms Pacman       &    307.3 &   6951.6 &    990.2 &   418.7 &    528.5 &   1620.1 &   293.3 &\textbf{   1847.1 }&   1715.9 &   1520.0   \\
Pong            &    -20.7 &     14.6 &     -4.4 &   -20.9 &    -20.4 &     -3.0 &   -21.0 &      0.9 &      1.7 &\textbf{      7.6 }  \\
Private Eye     &     24.9 &  69571.3 &     62.8 &   -20.7 &     89.4 &\textbf{    100.0 }&    12.7 &     98.2 &\textbf{    100.0 }&     90.0   \\
Qbert           &    163.9 &  13455.0 &\textbf{    720.0 }&   201.0 &    277.4 &    706.5 &   215.2 &    650.5 &    601.9 &    709.8   \\
Road Runner     &     11.5 &   7845.0 &   5428.4 &   780.3 &   5592.9 &  17698.4 &  2617.8 &  18229.4 &  17443.5 &\textbf{  18370.2 }  \\
Seaquest        &     68.4 &  42054.7 &    577.8 &   105.7 &    193.2 &    965.3 &   118.8 &\textbf{   1115.0 }&    792.1 &    728.4   \\
Up N Down       &    533.4 &  11693.2 &  46042.6 &   892.2 &   4399.7 &  58142.0 &  1313.4 &  52772.9 &  39771.3 &\textbf{  79228.8 }  \\ \midrule
Median HNS      &    0.000 &    1.000 &    0.343 &   0.009 &    0.060 &    0.411 &   0.029 &    0.512 &    0.629 &\textbf{    0.679 }  \\
Mean HNS        &    0.000 &    1.000 &    0.565 &  -0.054 &    0.181 &    0.943 &   0.098 &    1.004 &    0.978 &\textbf{    1.149 }  \\ \midrule
\#Games $>$ Human  &        0 &        0 &        3 &       0 &        1 &        7 &       0 &\textbf{        9 }&        8 &\textbf{        9 }  \\
\#Games $>$ SPR    &        0 &       19 &       10 &       1 &        1 &       18 &       1 &\textbf{       20 }&       19 &\textbf{       20 }  \\
\bottomrule
\end{tabular}

}

\end{table}

\clearpage
\section{Transferring Representations between Games}\label{app:transfer}
One advantage of pretraining representations is the possibility of representations being useful across games. Intuitively, we expect better transfer between similar games so we chose five ``cliques'' of games with similar semantics and visual elements.  The cliques are shown in Table~\ref{tab:cliques}.
We pretrain on a dataset of 750k frames from each game in a clique (i.e. 3M frames for a clique of 4) and finetune on a single game.
To show whether pretraining on other games is beneficial, we compare to a baseline of pretraining on just the 750k frames from the single Atari 100k game we use for finetuning. 

Our results in Table~\ref{tab:clique_scores} show that pretraining with the extra frames from the clique games is mostly unhelpful to finetune performance.
Only Kangaroo shows a modest improvement, a few games show no difference in performance, and most games show a decrease in performance when pretraining with other games.
We believe that Atari may not be as suitable to transferring representations as other domains, and previous work using Atari to learn transferable representations has also had negative results \citep{stooke2020decoupling}.
Though game semantics can be similar, we note that even small differences in rule sets and visual cues can make transfer difficult.

\begin{table}[ht]
\caption{Cliques of semantically similar games}
\label{tab:cliques}
\centering
\begin{tabular}{p{0.1\textwidth}p{0.4\textwidth}}
\toprule
Clique & Games \\
\midrule
space           & Space Invaders, Assault, Demon Attack, Phoenix \\
pacman          & MsPacman, Alien, Bank Heist, Wizard Of Wor \\      
platformer      & Montezuma Revenge, Hero, Kangaroo, Tutankham \\
top scroller    & Crazy Climber, Up N Down, Skiing, Journey Escape \\
side scroller   & Chopper Command, James Bond, Kung Fu Master, Private Eye  \\
\bottomrule
\end{tabular}
\end{table}
\begin{table}[ht]
\caption{
    Mean return per episode for clique games in Atari100k~\citep{simple} after 100k steps.
    Agents are evaluated at the end of training, and scores for all methods are averaged over 10 random seeds.
    Games in the same clique are placed together.
}
\label{tab:clique_scores}
\centering
\begin{tabular}{lrr}
\toprule
Game & Single & Clique \\
\midrule
Assault         &  738.5   & 554.1 \\
Demon Attack    &  1171.8  & 695.0 \\
\midrule
Alien           &  1183.9  & 830.2 \\
Bank Heist      &   448.8  & 303.0 \\
Ms Pacman       &  1595.8  & 1352.1 \\
\midrule
Kangaroo        &   489.2  & 994.0 \\
\midrule
Crazy Climber   &  52036.0 & 21829.8 \\ 
Up N Down       &  18974.7 &   13493.9 \\
\midrule
James Bond       &   397.6  & 325.4 \\
Kung Fu Master  & 16402.6  & 16499.0 \\
Chopper Command &   933.6  & 854.6 \\
\bottomrule
\end{tabular}
\end{table}

\end{document}